\documentclass[twoside]{article}
\usepackage[accepted]{aistats2016}
\usepackage{hyperref}
\usepackage{url}
\usepackage{graphicx}
\usepackage{amsfonts}
\usepackage{amsthm}
\usepackage{dsfont}
\usepackage{subcaption}
\usepackage{caption}
\usepackage{enumitem}
\usepackage{algorithm}
\usepackage[noend]{algpseudocode}
\usepackage{float}
\usepackage{pgfplots}
\usepackage{epstopdf}

\newcommand{\mycomment}[1]%
{\par {\bfseries \color{green} \scriptsize{#1} \par}}

\newcommand{\quiz}[1]%
{\par {\bfseries \color{red} \scriptsize{Quiz: #1} \par}} 
\newcommand{\review}[1]%
{\par {\bfseries \color{green} \scriptsize{#1} \par}}
\newcommand{\lab}[1]%
{\par {\bfseries \color{cyan} \scriptsize{#1} \par}}
\newcommand{\concept}[1]%
{\emph{\textbf{\color{blue} #1}}}

\newcommand\cut[1]{}

\usepackage{algorithm}

\newcommand{\squishlist}{
   \begin{list}{$\bullet$}
    { \setlength{\itemsep}{0pt}      \setlength{\parsep}{3pt}
      \setlength{\topsep}{3pt}       \setlength{\partopsep}{0pt}
      \setlength{\leftmargin}{1.5em} \setlength{\labelwidth}{1em}
      \setlength{\labelsep}{0.5em} } }

\newcommand{\squishlisttwo}{
   \begin{list}{$\bullet$}
    { \setlength{\itemsep}{0pt}    \setlength{\parsep}{0pt}
      \setlength{\topsep}{0pt}     \setlength{\partopsep}{0pt}
      \setlength{\leftmargin}{2em} \setlength{\labelwidth}{1.5em}
      \setlength{\labelsep}{0.5em} } }

\newcommand{\squishend}{
    \end{list}  }
 
\newtheorem{thm}{Theorem}[section]
\newtheorem{corr}{Corollary}[section]
\newtheorem{lemma}{Lemma}[section]

\newcommand{\real}{\mbox{$\mathbb{R}$}}

\newcommand{\tr}{\mbox{$\mbox{tr}$}}

\newcommand{\myexpect}{\mathbb{E}}

\newcommand{\myvec}[1]{\mbox{$\mathbf{#1}$}}
\newcommand{\myvecsym}[1]{\mbox{$\boldsymbol{#1}$}}

\newcommand{\vone}{\mbox{$\myvecsym{1}$}}

\newcommand{\va}{\mbox{$\myvec{a}$}}
\newcommand{\vb}{\mbox{$\myvec{b}$}}

\newcommand{\vd}{\mbox{$\myvec{d}$}}

\newcommand{\vf}{\mbox{$\myvec{f}$}}

\newcommand{\vk}{\mbox{$\myvec{k}$}}

\newcommand{\vx}{\mbox{$\myvec{x}$}}

\newcommand{\vy}{\mbox{$\myvec{y}$}}

\newcommand{\vA}{\mbox{$\myvec{A}$}}

\newcommand{\vD}{\mbox{$\myvec{D}$}}

\newcommand{\vI}{\mbox{$\myvec{I}$}}

\newcommand{\vK}{\mbox{$\myvec{K}$}}

\newcommand{\vM}{\mbox{$\myvec{M}$}}

\newcommand{\vU}{\mbox{$\myvec{U}$}}
\newcommand{\vV}{\mbox{$\myvec{V}$}}

\newcommand{\calA}{\mbox{${\cal A}$}}

\newcommand{\calI}{\mbox{${\cal I}$}}

\newcommand{\be}{\begin{equation}}
\newcommand{\ee}{\end{equation}}
\newcommand{\bea}{\begin{eqnarray}}
\newcommand{\eea}{\end{eqnarray}}
\newcommand{\beaa}{\begin{eqnarray*}}
\newcommand{\eeaa}{\end{eqnarray*}}

\usepackage{amsmath}

\DeclareMathOperator*{\argmax}{arg\,max}

\newcommand{\GP}{\mathcal{GP}}

\newcommand{\vKtil}{\widetilde{\mathbf{K}}}
\newcommand{\vktil}{\widetilde{\mathbf{k}}}
\newcommand{\gammaNTI}{\gamma_N}
\newcommand{\gammaNtilTI}{\gamma_{\tilde{N}}}
\newcommand{\gammaTTI}{\gamma_T}
\newcommand{\cov}{\mathrm{Cov}}

\newcommand{\Ltil}{\tilde{L}}
\newcommand{\ITI}{I}
\newcommand{\vertiii}[1]{{\left\vert\kern-0.25ex\left\vert\kern-0.25ex\left\vert #1 \right\vert\kern-0.25ex\right\vert\kern-0.25ex\right\vert}}
\newcommand{\Ndist}{\mathcal{N}}
\newcommand{\muTI}{\mu}
\newcommand{\sigmaTI}{\sigma}
\newcommand{\Ord}{\mathcal{O}}
\newcommand{\Otil}{\tilde{\mathcal{O}}}
\newcommand{\Itil}{\tilde{I}}
\newcommand{\gammatil}{\tilde{\gamma}}

\newcommand{\mutil}{\tilde{\mu}}
\newcommand{\sigmatil}{\tilde{\sigma}}
\newcommand{\Ntil}{\tilde{N}}
\newcommand{\balpha}{\boldsymbol{\alpha}}

\begin{document}

%

%

\twocolumn[

\aistatstitle{Time-Varying Gaussian Process Bandit Optimization}

\aistatsauthor{ Ilija Bogunovic, Jonathan Scarlett, Volkan Cevher }

\aistatsaddress{ 
    Laboratory for Information and Inference Systems (LIONS) \\ 
    \'Ecole Polytechnique F\'ed\'erale de Lausanne (EPFL) \\
    Email: \{ilija.bogunovic, jonathan.scarlett, volkan.cevher\}@epfl.ch, } 

]

\begin{abstract}
    We consider the sequential Bayesian optimization problem with bandit feedback, adopting a formulation that allows for the reward function to vary with time. We model the reward function using a Gaussian process whose evolution obeys a simple Markov model.  We introduce two natural extensions of the classical Gaussian process upper confidence bound (GP-UCB) algorithm. The first, R-GP-UCB, resets GP-UCB at regular intervals. The second, TV-GP-UCB, instead forgets about old data in a smooth fashion.  Our main contribution comprises of novel regret bounds for these algorithms, providing an explicit characterization of the trade-off between the time horizon and the rate at which the function varies.  We illustrate the performance of the algorithms on both synthetic and real data, and we find the gradual forgetting of TV-GP-UCB to perform favorably compared to the sharp resetting of R-GP-UCB.  Moreover, both algorithms significantly outperform classical GP-UCB, since it treats stale and fresh data equally.
\end{abstract}

\vspace*{-1ex}
\section{Introduction}
\vspace*{-1ex}

In recent years, there has been a great deal of interest in the theory and methods for bandit optimization problems, where one seeks to sequentially select a sequence of points to optimize an unknown reward function from noisy samples \cite{Sri12,Bub12,Lai85}.  Such problems have numerous applications, including sensors networks, recommender systems, and finance.  A key challenge is to rigorously trade-off between \emph{exploration}, i.e.,~learning the behavior of the function across the whole domain, and \emph{exploitation}, i.e.,~selecting points that have previously given high rewards.

In the vast majority of practical applications, the function to be optimized is not static, but varies with time: In sensor networks, measured quantities such as temperature undergo fluctuations; in recommender systems, the users' preferences may change according to external factors; similarly, financial markets are highly dynamic.  In such cases, the performance of standard algorithms may deteriorate, since these continue to treat stale data as being equally important as fresh data.   The development of algorithms and theory to handle time variations is therefore crucial.  

In this paper, we take a novel approach to handling time variations, modeling the reward function as a Gaussian process (GP) that varies according to a simple Markov model.

\textbf{Related Work:}
Even in time-invariant settings, GP-based models provide a flexible and powerful approach to Bayesian optimization problems \cite{Bro10,Sno12}.  Here, the smoothness properties of the reward function are dictated by a kernel function \cite{Ras06}.  A wide variety of works have made use of upper confidence bound (UCB) algorithms, where the selected point maximizes a linear combination of the posterior mean and standard deviation.  In particular, Srivinas \emph{et al.} \cite{Sri12} provided regret bounds for the GP-UCB algorithm, and several extensions were given subsequently, including the contextual \cite{Kra11} and high-dimensional \cite{Djo13,Sno14,Wan13} settings.

While the study of time-varying models is limited in the GP setting, several such models have been considered in the multi-armed bandit (MAB) setting.  Perhaps the most well-known one is the adversarial setting \cite{Lai85,Bub12,Bub15}, where one typically seeks to compete with the best fixed strategy.  Rewards modeled by Markov chains have been considered under the categories of \emph{restless bandits} \cite{Whi88,Ber00,Ort12,Sli08}, where the reward for each arm changes at each time step, and \emph{rested bandits} \cite{Tek12,Liu13}, where only the pulled arm changes. 


Two further related works are those of Slivkins and Upfal \cite{Sli08}, who studied a MAB problem with varying rewards based on Brownian motion, and  Besbes \emph{et al.}\cite{Bes14}, who considered a general MAB setting with time-varying rewards subject to a total budget in the amount of change allowed. Both \cite{Sli08} and \cite{Bes14} demonstrate the need for a \emph{forgetting-remembering trade-off} arising from the fact that using the information from more samples may decrease the variance of the function estimates, while older information may be stale and hence misleading.  Both papers present strategies in which the algorithm is reset at regular intervals in order to discard stale data.  This is shown to be optimal in the worst case for the function class considered in \cite{Bes14}, whereas in \cite{Sli08} it is shown that simple resetting strategies can be suboptimal in more specific scenarios, and alternative approaches are presented.


In contrast to GP-based settings such as ours, the setups of \cite{Sli08} and \cite{Bes14} consider finite action spaces, and assume independence between the rewards associated with different arms.  Thus, observing the reward of one arm does not reveal any information about the other ones, and the algorithms are designed to exploit temporal correlations, but not spatial correlations.

\textbf{Contributions:}
We introduce two algorithms for addressing the fundamental trade-offs inherent in the problem formulation: (i) trading off exploration with exploitation; (ii) differentiating between stale and fresh data in the presence of time variations; (iii) exploiting spatial and temporal correlations present in the reward function.  Our main results present regret bounds, first for general kernels and then for the squared-exponential and M\'atern kernels, that explicitly characterize the trade-off between the time horizon and the rate at which the function varies.   Their proofs require novel techniques to handle difficulties arising from the time variations, such as the maximum function value and its location changing drastically throughout the duration of the time horizon.  Moreover, we provide an algorithm-independent lower bound on the cumulative regret.  Finally, we demonstrate the utility of our model and algorithms on both synthetic and real-world data

\vspace*{-1ex}
\section{Problem Statement}
\vspace*{-1ex}

We seek to sequentially optimize an unknown reward function $f_t$ over a compact, convex subset $D \subset  \real^d$.\footnote{Finite domains were also handled in the time-invariant setting \cite{Sri12}, and all of our upper bounds have counterparts for such scenarios that are in fact simpler to obtain compared to the compact case.}  At time $t$, we can interact with $f_t$ only by querying at some point $x_t \in D$, after which we observe a noisy observation $y_t = f_t(x_t) + z_t$, where $z_t \sim \Ndist(0,\sigma^2)$. We assume that the noise realizations at different time instants are independent.  The goal is to maximize the reward via a suitable trade-off between exploration and exploitation.  This problem is ill-posed for arbitrary reward functions even in the time-invariant setting, and it is thus necessary to introduce suitable smoothness assumptions.  We take the approach of \cite{Sri12}, and model the reward function as a sample from a Gaussian process, where its smoothness is dictated by the choice of kernel function.

\textbf{Model for the Reward Functions:}
Let $k : D \times D \to \real_+$ be a kernel function, and let $\GP(\mu,k)$ be a Gaussian process \cite{Ras06} with mean $\mu \in \real^d$ and kernel $k$.  As in \cite{Sri12}, we assume bounded variance: $\forall x \in D, k(x,x) \leq 1$. Two common kernels are squared exponential (SE) and Mat\'ern, defined as
\begin{align}
    k_{\text{SE}}(x,x') &= \exp \bigg(- \dfrac{\|x - x'\|^2}{2l^2} \bigg) \\ 
    k_{\text{Mat\'ern}}(x,x') &= \dfrac{2^{1-\nu}}{\Gamma(\nu)} \bigg(\dfrac{\sqrt{2\nu}\|x - x'\|}{l}\bigg)^{\nu} \nonumber \\
        &\qquad\qquad \times B_{\nu}\bigg(\dfrac{\sqrt{2 \nu}\|x - x'\|}{l} \bigg),
    \end{align}
where $l>0$ and $\nu > 0$ are hyperparameters,  and $B_{\nu}$ denotes the modified Bessel function.

\begin{figure*}
\centering
        \begin{subfigure}[b]{0.33\textwidth}
                 \includegraphics[width= \columnwidth]{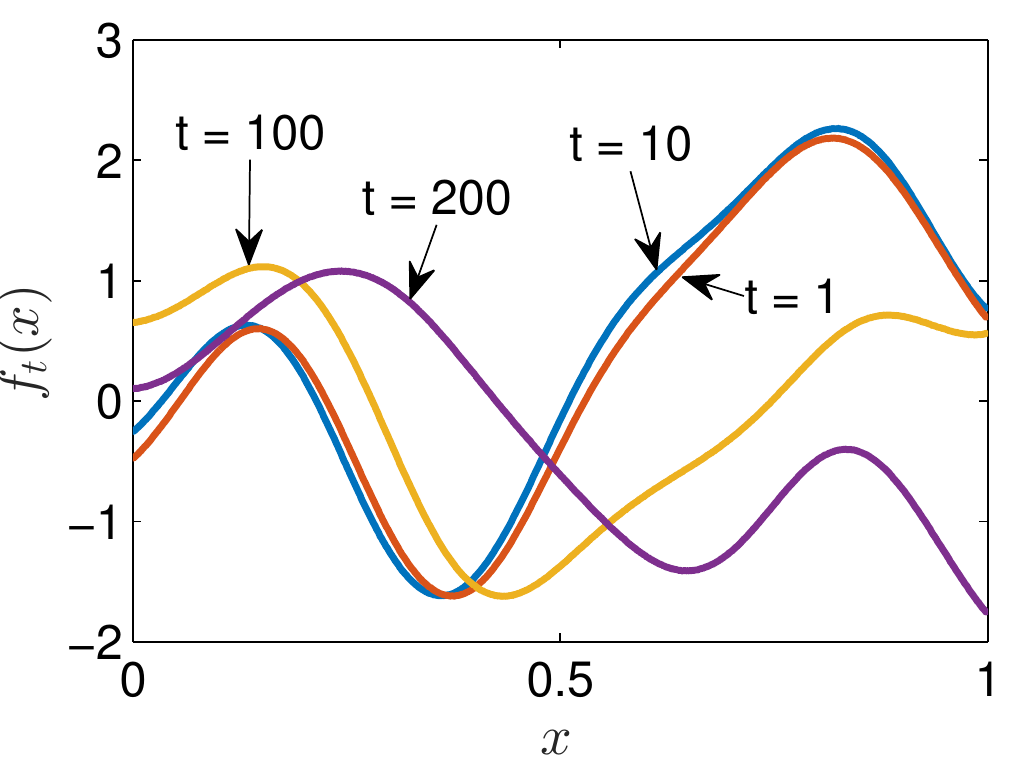}
                 \par
        \end{subfigure}
        \hspace{10mm}
        \begin{subfigure}[b]{0.33\textwidth}
                 \includegraphics[width=\columnwidth]{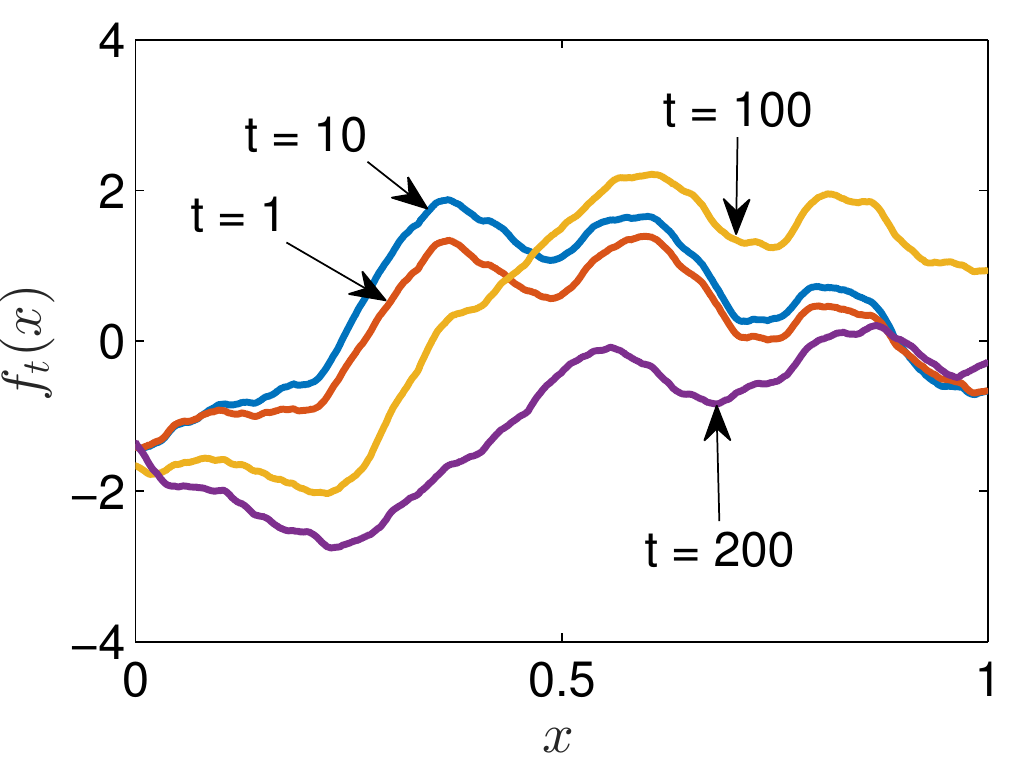}
                 \par
        \end{subfigure}
        \caption{Two examples of GP functions when $\epsilon = 0.01$: (Left) Squared exponential kernel ($l = 0.2$); (Right) Mat\'ern kernel ($l = 0.2$, $\nu = 1.5$).  Note that the location of the maximum changes significantly at distant times. } \label{fig:example}
\end{figure*}

Letting $g_1,g_2,\dotsc$ be independent random functions on $D$ with $g_i \sim \GP(0,k)$, the reward functions are modeled as follows:
\begin{gather}
        f_1(x) = g_1(x) \label{eq:tv_model0} \\
        f_{t+1}(x) = \sqrt{1 - \epsilon}\,f_{t}(x) +  \sqrt{\epsilon}\,g_{t+1}(x) \quad \forall t \ge 2, \label{eq:tv_model}
\end{gather}
where $\epsilon \in [0,1]$ quantifies how much the function changes after every time step. If $\epsilon = 0$ then we recover the standard time-invariant model \cite{Sri12}, whereas if $\epsilon = 1$ then the reward functions are independent between time steps.  Importantly, for any choice of $\epsilon$ we have for all $t$ that $f_t \sim \GP(0,k)$.  See Figure \ref{fig:example} for an illustration.

From a practical perspective, this model has the desirable property of only having one additional hyperparameter $\epsilon$ compared to the standard GP model, thus facilitating the learning process. It serves as a suitable model for reward functions that vary at a steady rate, though we will see numerically in Section \ref{sec:NUMERICAL} that the resulting algorithms are also effective more generally. 

As noted in \emph{regression} studies in \cite{Van12,Van12a}, our model is equivalent to a spatiotemporal kernel model with temporal kernel $(1-\epsilon)^{|t_1-t_2|/2}$.  We expect our techniques to apply similarly to other temporal kernels, particularly \emph{stationary} kernel functions that depend only on the time difference $|t_1 - t_2|$, but we focus on \eqref{eq:tv_model0}--\eqref{eq:tv_model} for concreteness. Spatiotemporal kernels can also be considered in the contextual bandit setting \cite{Kra11}, but to our knowledge, no regret bounds have been given that explicitly characterize the dependence on the function's rate of variation, as is done in our main result.


\textbf{Regret:}
Let $x^*_t$ denote a maximizer of $f_t$ at time $t$, i.e.,~$x_t^* = \argmax_{x} f_t(x)$, and suppose that our choice at time $t$ is $x_t$. Then the \textit{instantaneous regret} we incur at time $t$ is $r_t = f_t(x^*_t) - f_t(x_t)$.  We are interested in minimizing the \textit{cumulative regret} $R_T = \sum_{t=1}^T r_t$.   

These definitions naturally coincide with those for the time-invariant setting when $\epsilon = 0$.  Note that we do not aim to merely compete with fixed strategies, but instead to track the maximum of $f_t$ for all $t$.  In our setting, a notion of regret based on competing with a fixed strategy would typically lead to a negative cumulative regret. In other words, \emph{all fixed strategies perform poorly}.


In time-invariant scenarios, as well as several time-varying scenarios, algorithms are typically designed to achieve \emph{sublinear regret}.  In our setting, we will show that for \emph{fixed} $\epsilon$, the cumulative regret $R_T$ must in fact be $\Omega(T)$ (\emph{cf.},~Theorem \ref{thm:conv}).  Intuitively, this is because if the function changes significantly at each time step, one cannot expect to track its maximum to arbitrary precision.  However, we emphasize that what is really of interest is the \emph{joint} dependence of $R_T$ on $T$ and $\epsilon$, and we thus seek regret bounds of the form $\Otil(T\psi(\epsilon))$ for some function $\psi(\epsilon)$ that vanishes as $\epsilon \to 0$.\footnote{Here and subsequently, the notation $\Otil(\cdot)$ denotes asymptotics up to logarithmic factors.}  Our approach is analogous to  Slivkins and Upfal \cite{Sli08}, who considered another time-varying setting with unavoidable $\Omega(T)$ regret for any fixed function variation parameter, and focused on the behavior in the implied constant in the limit as that parameter vanishes. 

For the squared exponential and Mat\'ern kernels, we obtain regret bounds of the form $\Otil(T\epsilon^{\alpha})$ for some $\alpha > 0$ (\emph{cf.},~Corollary \ref{cor:specific}), which can be viewed as being sublinear whenever $\epsilon = \Ord(T^{-c})$ for some $c > 0$.  We observe that when $c < 1$, the correlation between $f_1(x)$ and $f_T(x)$ is negligible, meaning that the corresponding maximum may (and typically will) change drastically over the duration of the time horizon, e.g.,~see Figure \ref{fig:example}.  


\textbf{Limitations of GP-UCB:}
We briefly recall the GP-UCB algorithm from \cite{Sri12}, in which at each time step the selected point maximizes a function of the form $\muTI_{t-1}(x) + \sqrt{\beta_t}\sigmaTI_{t-1}(x)$.  Here, defining $\vK_t = \big[k(x,x')\big]_{x,x' \in \vx_t}$ and $\vk_t(x) = \big[k(x_i,x)\big]_{i=1}^t$, the quantities
\begin{gather}
   	\muTI_{t+1}(x)  :=  \vk_t(x)^T\big(\vK_t + \sigma^2\vI_t \big)^{-1} \vy_t \label{eq:mu_update_TI} \\
   	\sigmaTI_{t+1}(x,x)^2 := k(x,x) - \vk_t(x)^T \big(\vK_t + \sigma^2\vI_t \big)^{-1} \vk_t(x), \label{eq:sigma_update_TI}  
\end{gather}
are the posterior mean and variance of the time-invariant GP $f(x)$, respectively, given the previous samples $\vx_t = [x_1,\dotsc,x_t]$ and corresponding observations $y_1,\dotsc,y_t$.  Intuitively, one seeks points with a high mean $\muTI_t$ to favor exploitation, but with a high standard deviation $\sigmaTI_t$ to favor exploration.

In the time-invariant setting, GP-UCB is known to achieve sublinear regret under mild assumptions \cite{Sri12}.  As mentioned above, the problem with using it in our setting is that it treats all of the previous samples as being equally important, whereas according to our model, the samples become increasingly stale with time.  We now proceed to describing our algorithms that account for this fact.

\vspace*{-1ex}
\section{Algorithms} \label{sec:ALGO}
\vspace*{-1ex}

We first introduce an algorithm R-GP-UCB that takes a conceptually simple approach to handling the forgetting-remembering trade-off, namely, running the GP-UCB algorithm within blocks of size $N$, and applying resetting at the start of each block.  Some insight on how to choose $N$ is given by our bounds in the following section. The pseudo-code is shown in Algorithm \ref{alg:R}.

\begin{algorithm}
    \caption{GP-UCB with Resetting (R-GP-UCB)} \label{r-gp-ucb}
    \begin{algorithmic}[1]
    \Require Domain $D$, GP prior ($\mu_0$, $\sigma_0$, $k$), block size $N$
    \For {$t = 1,2 ...$}
    	\If{$t~\mathrm{mod}~N = 1$}
        	\State Reset $\muTI_{t-1}(x) = \mu_0(x)$ and $\sigmaTI_{t-1}(x) = \sigma_0(x)$ 
        	\State for each $x$
    	\EndIf
    	\State Choose $x_t = \argmax_{x \in D} \muTI_{t-1}(x) + \sqrt{\beta_t} \sigmaTI_{t-1}(x)$
    	\State Sample $y_t = f_t(x_t) + z_t$
    	\State Perform Bayesian update as in \eqref{eq:mu_update_TI}--\eqref{eq:sigma_update_TI}, using 
    	\State only the samples $\{x_t\}$ and $\{y_t\}$ obtained since 
    	\State the most recent reset, to obtain $\muTI_t$ and $\sigmaTI_t$
    \EndFor
    \end{algorithmic} \label{alg:R}
\end{algorithm}

Our second algorithm, TV-GP-UCB, instead forgets in a ``smooth'' fashion, by using a posterior update rule obtained via the time-varying model \eqref{eq:tv_model0}--\eqref{eq:tv_model}.  In analogy with \eqref{eq:mu_update_TI}--\eqref{eq:sigma_update_TI}, the mean and variance of $f_t$ given the previous samples $\vx_t = (x_1,\dotsc,x_t)$ and corresponding observations $y_1,\dotsc,y_t$ are given by
\begin{gather}
   	\mutil_{t+1} (x)  :=  \vktil_t(x)^T\big(\vKtil_t + \sigma^2\vI_t \big)^{-1} \vy_t \label{eq:mu_update} \\
   	\sigmatil^2_{t+1}(x,x') := k(x,x) - \vktil_t(x)^T \big(\vKtil_t + \sigma^2\vI_t \big)^{-1} \vktil_t(x), \label{eq:sigma_update} 
\end{gather}
where $\vKtil_t = \vK_t \circ \vD_t$ with $\vD_t = \big[ \left(1- \epsilon\right)^{|i-j|/2}\big]_{i,j =1}^T$, and $\vktil_t(x) = \vk_t(x) \circ \vd_t$ with $\vd_t = \big[ \left(1- \epsilon\right)^{(T+1-i)/2}\big]_{i =1}^T$.  Here $\circ$ is the Hadamard product, and $\vI_k$ is the $k \times k$ identity matrix. 

The derivation of \eqref{eq:mu_update}--\eqref{eq:sigma_update} is given in the supplementary material.  Using these updates, the pseudo-code for the TV-GP-UCB algorithm is given in Algorithm \ref{alg:TV}.  The idea is that the older a sample is, the smaller the value in the corresponding entries of $\vd_t$ and $\vD_t$ defined following \eqref{eq:sigma_update}, and hence the less it contributes to the final values of $\mutil_t(x)$ and $\sigmatil_t(x)$.  This algorithm can in fact be considered a special case of contextual GP-UCB \cite{Kra11} with a spatio-temporal kernel, but our analysis (Section \ref{sec:BOUNDS}) goes far beyond that of \cite{Kra11} in order to explicitly characterize the dependence on $T$ and $\epsilon$.

\begin{algorithm}
    \caption{Time-Varying GP-UCB (TV-GP-UCB)} \label{tv-gp-ucb}
    \begin{algorithmic}[1]
    \Require Domain $D$, GP prior ($\tilde{\mu}_0$, $\tilde{\sigma}_0$, $k$) and parameter $\epsilon$
    \For {$t = 1,2 ...$}
    	\State Choose $x_t = \argmax_{x \in D} \mutil_{t-1}(x) + \sqrt{\beta_t} \sigmatil_{t-1}(x)$
    	\State Sample $y_t = f_t(x_t) + z_t$
    	\State Perform Bayesian update as in \eqref{eq:mu_update}--\eqref{eq:sigma_update} to obtain $\mutil_t$ and $\sigmatil_t$
    \EndFor
    \end{algorithmic} \label{alg:TV}
\end{algorithm}

\textbf{Computational Complexity:} As it is presented above, TV-GP-UCB has an identical computational complexity to GP-UCB, i.e. the complexity of the sequential Bayesian update is $\mathcal{O}(T^2)$ ~\cite{Osb08}.  R-GP-UCB is less complex, since the matrix operations are on matrices of size $N$ rather than the overall time horizon $T$.  In practice, however, one could further modify TV-GP-UCB to improve the efficiency by occasionally resetting and/or discarding stale data \cite{Osb08}.

\vspace*{-1ex}
\section{Theoretical Bounds} \label{sec:BOUNDS}
\vspace*{-1ex}

In this section, we provide our main theoretical upper and lower bounds on the regret. We assume throughout this section that hyperparameters are known, i.e. both spatial kernel hyperparameters and $\epsilon$; in the numerical section (Section \ref{sec:NUMERICAL}) we will address real-world problems where these are unknown.  All proofs are given in the supplementary material.

\subsection{Preliminary Definitions and Results}

\textbf{Smoothness Assumptions:}
Each of our results below will assume that the kernel $k$ is such that a (strict) subset of the following statements hold for some $(a_i,b_i)$ and all $L \ge 0$:
\begin{align}
    \Pr\bigg( \sup_{x \in D} \big|f(x)\big| > L \bigg) &\le a_0 e^{-(L/b_0)^2} \label{eq:k_deriv0} \\
    \Pr\bigg( \sup_{x \in D} \Big|\frac{\partial f}{\partial x^{(j)}}\Big| > L \bigg) &\le a_1 e^{-(L/b_1)^2}, \nonumber \\
        & \quad~~ j=1,\dotsc,d \label{eq:k_deriv1} \\
    Pr\bigg( \sup_{x \in D} \Big|\frac{\partial^2 f}{\partial x^{(j_1)}\partial x^{(j_2)}}\Big| > L \bigg) &\le a_2 e^{-(L/b_2)^2}, \nonumber \\
        &  j_1,j_2=1,\dotsc,d, \label{eq:k_deriv2}
\end{align}
where $f \sim \GP(0,k)$.  Assumption \eqref{eq:k_deriv0} is mild, since $f(x)$ is Gaussian and thus has exponential tails.  Assumption \eqref{eq:k_deriv1} was used in \cite{Sri12}, and ensures that the behavior of the GP is not too erratic.  It is satisfied for the SE kernel, as well as the Mat\'ern kernel with $\nu > 2$ \cite{Sri12}, though for other kernels (e.g.,~Ornstein-Uhlenbeck) it can fail. Assumption \eqref{eq:k_deriv2} is used only for our lower bound; it is again satisfied by the SE kernel, as well as the Mat\'ern kernel with $\nu > 4$.

\textbf{Mutual Information:}
It was shown in \cite{Sri12} that a key quantity governing the regret bounds of GP-UCB in the time-invariant setting is the mutual information
\begin{equation}
    \ITI(\vf_T;\vy_T) = \frac{1}{2}\log\det\big( \vI_T + \sigma^{-2} \vK_T \big), \label{eq:MI_TI} 
\end{equation}
where $\vf_T := \vf_T(\vx_T) = (f(x_1),\dotsc,f(x_T))$ for the time-invariant GP $f$. The corresponding maximum over any set of points $\vx_T=(x_1,\dotsc,x_T)$ is given by
\begin{equation}
    \gammaTTI :=  \max_{x_1,\dotsc,x_T} \ITI(\vf_T;\vy_T). \label{eq:gamma_TI}
\end{equation}
In our setting, the analogous quantities are as follows:
\begin{gather}
    \Itil(\vf_T;\vy_T) = \frac{1}{2}\log\det\big( \vI_T + \sigma^{-2} \vKtil_T \big), \label{eq:MI} \\
    \gammatil_T := \max_{x_1,\dotsc,x_T} \Itil(\vf_T;\vy_T), \label{eq:gamma}
\end{gather}
where $\vf_T := \vf_T(\vx_T) = (f_1(x_1),\dotsc,f_T(x_T))$.  While these take the same form as \eqref{eq:MI_TI}--\eqref{eq:gamma_TI}, they can behave significantly differently when $\epsilon > 0$. In particular, the time-varying versions are typically much higher due to the fact $\vf_T$ represents the points of $T$ different random functions, as opposed to a single function at $T$ different points.


\textbf{Algorithm-Independent Lower Bound:}
The following result gives an asymptotic lower bound for any bandit optimization algorithm under fairly mild assumptions, expressed in terms of the time horizon $T$ and parameter $\epsilon$.   
\begin{thm} \label{thm:conv}
    Suppose that the kernel is such that $f \sim \GP(0,k)$ is almost surely twice continuously differentiable and satisfies \eqref{eq:k_deriv1}--\eqref{eq:k_deriv2} for some $(a_1,b_1,a_2,b_2)$.  Then, any GP bandit optimization algorithm incurs expected regret $\myexpect[R_T] = \Omega(T\epsilon)$.
\end{thm}

The proof reveals that this result holds true even in the full information (as opposed to bandit) setting, and is based on the fact that at each time step, there is a non-zero probability that the maximum value and its location change by an amount proportional to $\epsilon$.  As discussed above, this lower bound motivates the study of the \emph{joint} dependence on the regret of $T$ and $\epsilon$, and in particular, the highest possible constant $\alpha$ such that the regret behaves as $\Otil(T\epsilon^{\alpha})$.


\subsection{Main Results}

We now present our main general bounds on the algorithms introduced in Section \ref{sec:ALGO}.  The two provide regret bounds of a similar form, but we will shortly apply these to specific kernels and find that the bounds for TV-GP-UCB yield better scaling laws.

\textbf{General Regret Bounds:}
The following theorems provide regret bounds for R-GP-UCB and TV-GP-UCB, respectively.  We will simplify these bounds below to obtain scaling laws for specific kernels.

\begin{thm} \label{thm:general_R}
    Let the domain $D \subset [0,r]^d$ be compact and convex, and suppose that the kernel is such that $f \sim \GP(0,k)$ is almost surely continuously differentiable and satisfies \eqref{eq:k_deriv0}--\eqref{eq:k_deriv1} for some $(a_0,b_0,a_1,b_1)$. 
    Fix $\delta \in (0,1)$, and set     
   \begin{equation}
       \beta_T = 2 \log \frac{2\pi^2 T^2}{3\delta} + 2d \log\bigg(rdbT^2\sqrt{\log\frac{2da \pi^2 T^2}{3\delta}}\bigg). \label{eq:beta_T_R}
   \end{equation}
    Defining $C_1 = 8/\log(1+\sigma^{-2})$, the R-GP-UCB algorithm satisfies the following after $T$ time steps:
    \begin{equation}
        R_T \le \sqrt{C_1 T \beta_T \bigg(\frac{T}{N} + 1\bigg) \gammaNTI} + 2 + T\psi_T(N,\epsilon) \label{eq:general_bound_R}
    \end{equation}
    with probability at least $1-\delta$, where
    \begin{multline}
        \psi_T(N,\epsilon) := \sqrt{\beta_T\big( 3\sigma^{-2} + \sigma^{-4} \big) N^3 \epsilon} \\ + \big(\sigma^{-2} + \sigma^{-4}\big) N^3 \epsilon (2+b_0) \sqrt{\log \frac{2(1+a_0) \pi^2 T^2}{3\delta}}.
    \end{multline}
\end{thm}

The proof of Theorem \ref{thm:general_R} departs from regular Bayesian optimization proofs such as \cite{Sri12} in the sense that the posterior updates \eqref{eq:mu_update_TI}--\eqref{eq:sigma_update_TI} assumed by the algorithm differ from the true posterior described by \eqref{eq:mu_update}--\eqref{eq:sigma_update}, thus requiring a careful handling of the effect of the mismatch.

\begin{thm} \label{thm:general}
    Let the domain $D \subset [0,r]^d$ be compact and convex, and suppose that the kernel is such that $f \sim \GP(0,k)$ is almost surely continuously differentiable and satisfies \eqref{eq:k_deriv1} for some $(a_1,b_1)$.   Fix $\delta \in (0,1)$, and set
    \begin{equation}
    	\beta_T = 2 \log \frac{\pi^2 T^2}{2\delta} + 2d \log\bigg(rdbT^2\sqrt{\log\frac{da \pi^2 T^2}{2\delta}}\bigg). \label{eq:beta_T}
    \end{equation}
    Defining $C_1 = 8/\log(1+\sigma^{-2})$, the TV-GP-UCB algorithm satisfies the following after $T$ time steps:
    \begin{align}
        R_T &\le \sqrt{C_1 T \beta_T \gammatil_T} + 2 \label{eq:general_bound} \\
               &\le \sqrt{C_1 T \beta_T \bigg(\frac{T}{\Ntil} + 1\bigg) \Big(\gammaNtilTI + \Ntil^3 \epsilon\Big)} + 2 \label{eq:weakened2}
    \end{align}
    with probability at least $1-\delta$, where \eqref{eq:weakened2} holds for any $\Ntil \in \{1,\dotsc,T\}$.
\end{thm}
\begin{figure*}[t!]
\centering
\setcounter{subfigure}{0}
  \begin{subfigure}[b]{0.24\textwidth}
  	 \includegraphics[width=\textwidth]{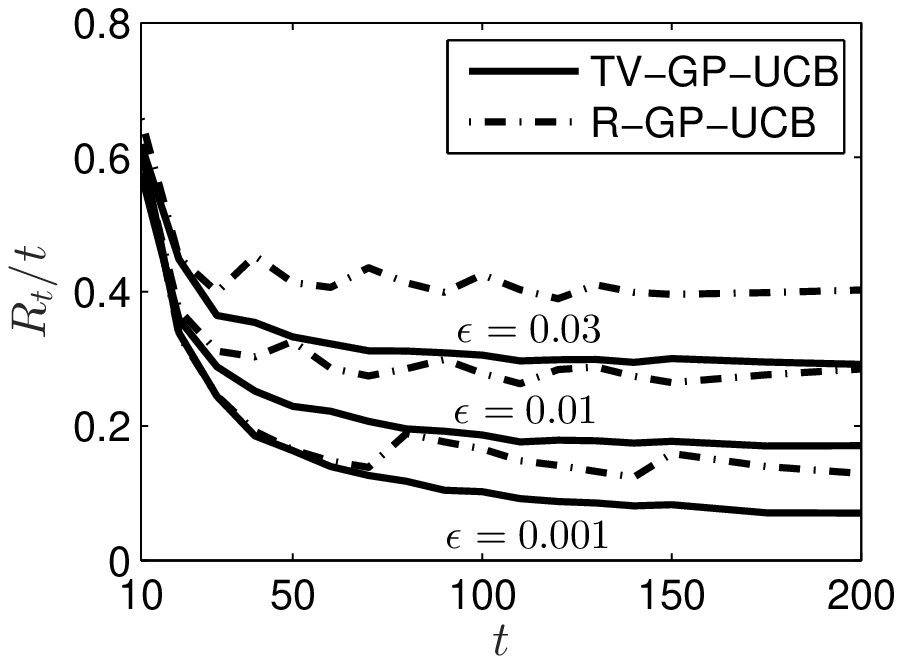}
  	 \caption{SE kernel}
  	 \label{fig: SE}
  \end{subfigure}
\setcounter{subfigure}{1}
  \begin{subfigure}[b]{0.24\textwidth}
    \includegraphics[width=\textwidth]{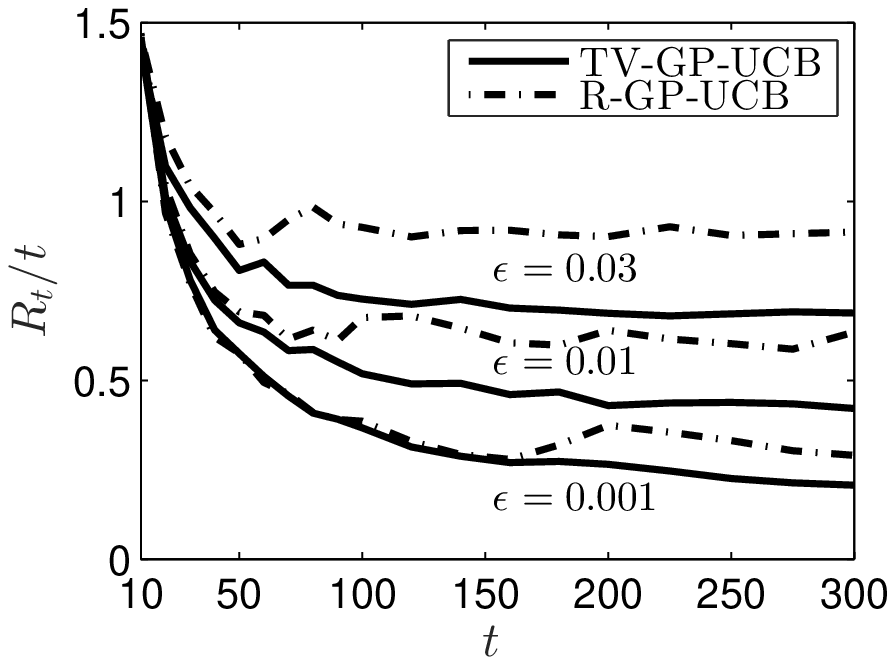}
	\caption{Mat\'ern52 kernel} 
	\label{fig:MK}
  \end{subfigure}
\setcounter{subfigure}{2}
	\begin{subfigure}[b]{0.24\textwidth}
    \includegraphics[width=\textwidth]{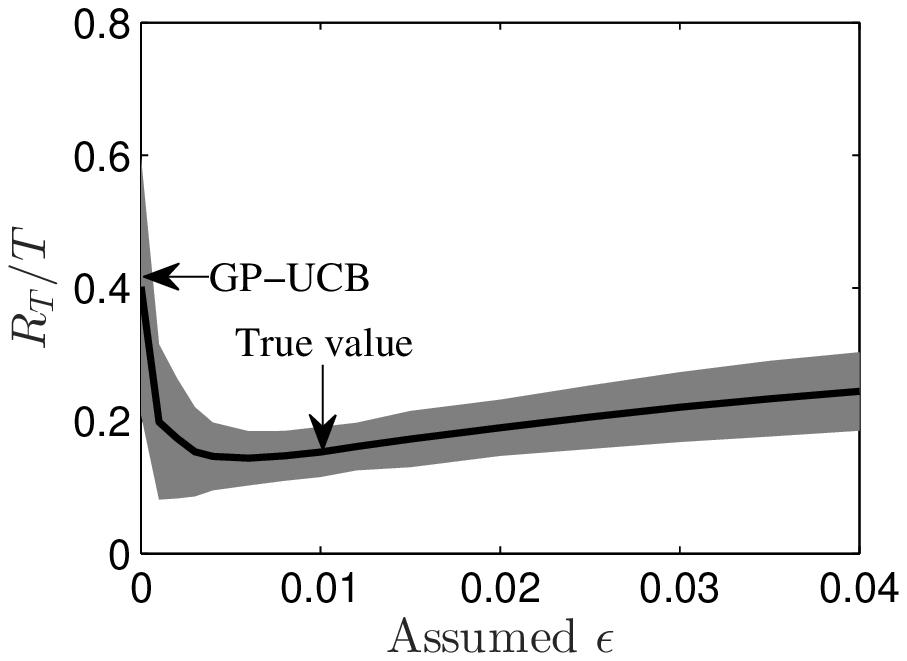}
	\caption{Mismatched TV-UCB} 
	\label{fig:mis_eps}
  \end{subfigure}
  \setcounter{subfigure}{3}
  \begin{subfigure}[b]{0.24\textwidth}
      \includegraphics[width=\textwidth]{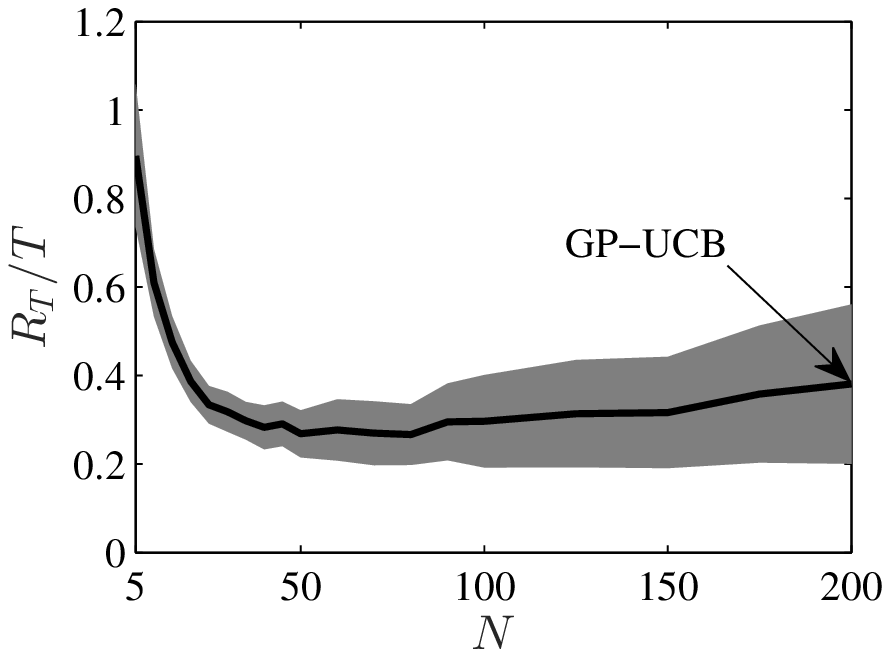}
	  \caption{Mismatched R-GP-UCB}
	\label{fig:block_size} 
  \end{subfigure}
   \caption{Numerical performance of upper confidence bound algorithms on synthetic data.}
\end{figure*}

The step in \eqref{eq:general_bound} is obtained using techniques similar to those of \cite{Sri12,Kra11}, whereas the step in \eqref{eq:weakened2} is non-trivial and new.  This step is key to our analysis, bounding the maximum mutual information $\gammatil_T$ for the time varying case in terms of the analogous quantity $\gammaNtilTI$ from the time-invariant setting.  The idea in doing this is to split the block $\{1,\dotsc,T\}$ into smaller blocks of size $\Ntil$ within which the overall variation in $f_t$ is not too large.  This is in contrast with R-GP-UCB (and \cite{Bes14}), where the algorithm takes the block length $N$ as a parameter and explicitly resets the algorithm every $N$ time steps.  For TV-GP-UCB, the length $\Ntil$ is only introduced as a tool in the analysis.

\textbf{Applications to Specific Kernels:}
Specializing the above results to the squared exponential and Mat\'ern kernels, using the corresponding bounds on $\gammaNTI$ from \cite{Sri12}, and optimizing $N$ as a function of $T$ and $\epsilon$, we obtain the following. 

\begin{corr} \label{cor:specific}
    Under the conditions of Theorems \ref{thm:general_R} and \ref{thm:general}, we have the following for any fixed $d$:
    \begin{enumerate}
        \item For the squared exponential kernel, $R_T = \Otil(\max\{\sqrt{T},T\epsilon^{1/8}\} )$  for R-GP-UCB with $N=\Theta(\min\{T,\epsilon^{-1/4}\})$, and $R_T = \Otil(\max\{\sqrt{T},T\epsilon^{1/6}\} )$ for TV-GP-UCB.
        \item Consider the Mat\'ern kernel with parameter $\nu > 2$, and set $c=\frac{d(d+1)}{2\nu + d(d+1)} \in (0,1)$.  We have $R_T = \Otil(\max\{\sqrt{T^{1+c}}, T\epsilon^{\frac{1}{2} \frac{1-c}{4-c} } \} )$ for R-GP-UCB with $N=\Theta(\min\{T,\epsilon^{-\frac{1}{4-c}}\})$, and $R_T = \Otil(\max\{\sqrt{T^{1+c}},T\epsilon^{\frac{1}{2} \frac{1-c}{3-c} } \} )$ for TV-GP-UCB.
    \end{enumerate}
\end{corr}

Observe that, upon substituting $\epsilon = 0$, the preceding $\Otil(\cdot)$ terms are dominated by the first terms in the maxima, and the bounds for both algorithms reduce to those in \cite{Sri12}.  In the case that $\epsilon$ vanishes more slowly (e.g.,~as $1/\sqrt{T}$), the regret bounds for TV-GP-UCB are strictly better than those of R-GP-UCB.  The worsened bounds for the latter arise due to the above-mentioned mismatch in the update rules.

For both kernels, the optimized block length $N$ of R-GP-UCB increases as $\epsilon$ decreases; this is to be expected, as it means that older samples are more correlated with the present function.  We also observe that $N$ increases as the function becomes smoother (by increasing $\nu$ for the Mat\'ern kernel, or by switching from Mat\'ern to squared exponential).

\vspace*{-1ex}
\section{Experiments} \label{sec:NUMERICAL}
\vspace*{-1ex}

In this section, we test our algorithms on both synthetic and real data, as well as studying the effect of mismatch with respect to the algorithm parameters $\epsilon$ and $N$.


\textbf{Practical considerations:} While \eqref{eq:beta_T_R} and \eqref{eq:beta_T} give explicit choices for $\beta_t$, these usually tend to be too conservative in practice. For good empirical performance, we rely only on the \emph{scaling} $\beta_t = O(d\log t)$ dictated by these choices, letting $\beta_t = c_1 \log(c_2 t)$ (similarly to \cite{Sri12,Kan15}, for example). We found $c_1 = 0.8$ and $c_2 = 4$ to be suitable for trading off exploration and exploitation, and we therefore use these in all of our synthetic experiments.

Our theoretical analysis assumed that we know the hyperparameters of both spatial and temporal kernel. Having perfect knowledge of $\epsilon$ and other hyperparameters is typically not realistic. The GP perspective allows us to select them in a principled way by maximizing the GP marginal likelihood \cite{Ras06}. In our real-world experiments below, we select $\epsilon$ in such manner, using the approach from \cite{Van12}, outlined in Appendix \ref{sec:OPT_EPS}.  In our synthetic experiments, we consider both the cases of perfect and imperfect knowledge of $\epsilon$.

\textbf{Baseline Comparisons:} We are not aware of any algorithms other than those in Section \ref{sec:ALGO} that exploit both spatial and temporal correlations.  In both our synthetic and real-data experiments, we found it crucial to handle both of these in order to obtain reasonable values for the cumulative regret, thus drastically limiting the number of reasonable baselines.  Nevertheless, we also consider GP-UCB (which exploits spatial but not temporal correlations), and in the real-world experiments, we consider a completely random selection (thus corresponding to a choice that we should hope to beat significantly).

\subsection{Synthetic Data}

We consider a two-dimensional setting and quantize the decision space  $D = [0,1]^2$ into $50 \times 50$ equally-spaced points.  We generate our data according to the time-varying model \eqref{eq:tv_model}, considering both the squared exponential and Mat\'ern kernels.  For the former we set $l = 0.2$, and for the latter we set  $\nu = 2.5$ and $l = 0.2$.  We set the sampling noise variance $\sigma^2$ to $0.01$, i.e. $1$\% of the signal variance. 

\textbf{Matched Case:} We first consider the case that the algorithm parameters are ``matched''.  Specifically, the parameter $\epsilon$ for TV-GP-UCB is the true parameter for the model, and the parameter $N$ for R-GP-UCB is chosen in accordance with Corollary \ref{cor:specific}:  $N=\lceil\min\big\{ T, 12 \epsilon^{-1/4} \big\}\rceil$ for the squared exponential kernel, and $N = \lceil\min \big\{ T, 24 \epsilon^{-\frac{1}{4-c}} \big\}\rceil$ for the Mat\'ern kernel, where the constants were found via cross-validation.

In Figures~\ref{fig: SE} and~\ref{fig:MK}, we show the average regret $\frac{R_T}{T}$ of TV-GP-UCB and R-GP-UCB for $\epsilon \in \lbrace 0.001, 0.01, 0.03 \rbrace$.  For each time shown, we average the performance over $200$ independent trials. We observe that for all values of $\epsilon$ and for both kernel functions, TV-GP-UCB outperforms R-GP-UCB, which is consistent with the theoretical bounds we obtained in the previous section.  Furthermore, we see that the curves for R-GP-UCB have an oscillatory behavior, resulting from the fact that one tends to incur more regret just after a reset is done.  In contrast, the curves for TV-GP-UCB are more steady, since the algorithm forgets in a ``smooth'' fashion.

\textbf{Mismatch and Robustness:}
We consider the stability of TV-GP-UCB when there is mismatch between the true $\epsilon$ and the one used in TV-GP-UCB. We focus on the squared exponential kernel, and we set $\epsilon = 0.01$ and $T = 200$. From Figure \ref{fig:mis_eps}, we see that the performance of TV-GP-UCB is robust with respect to the mis-specification of $\epsilon$. In particular, the increase in regret as $\epsilon$ is increasingly over-estimated is slow.  In contrast, while slightly under-estimating $\epsilon$ is not harmful, the regret increases rapidly beyond a certain point.  In particular, using $0$ instead of the true $\epsilon$ corresponds to simply running the standard GP-UCB algorithm, and gives the worst performance within the range shown.  Note that the shaded area corresponds to a standard deviation from the mean.

Next, we study R-GP-UCB on the same model to determine the robustness with respect to the choice of $N$; the results are shown in Figure \ref{fig:block_size}.  Values of $N$ that are too small are problematic, since the algorithm resets too frequently.  While the \emph{mean} of the regret is robust with respect to increasing $N$, we observe that the corresponding standard deviation also steadily increases.  GP-UCB is again recovered as a special case, corresponding to $N = T$.

\begin{figure*}[t!]
\centering
\setcounter{subfigure}{0}
 \begin{subfigure}[b]{0.3\textwidth}
	\includegraphics[width=\textwidth]{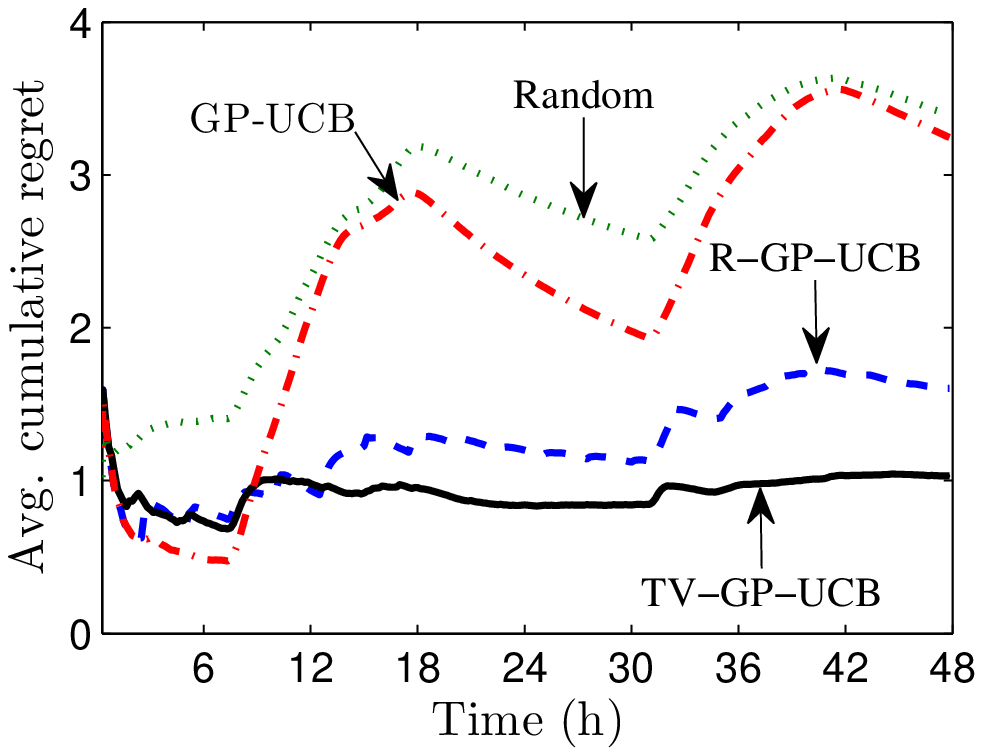}  
	\caption{Temp. data performance}
	\label{fig:real_data}
  \end{subfigure}
\setcounter{subfigure}{2}
  \begin{subfigure}[b]{0.3\textwidth}
    \includegraphics[width=\textwidth]{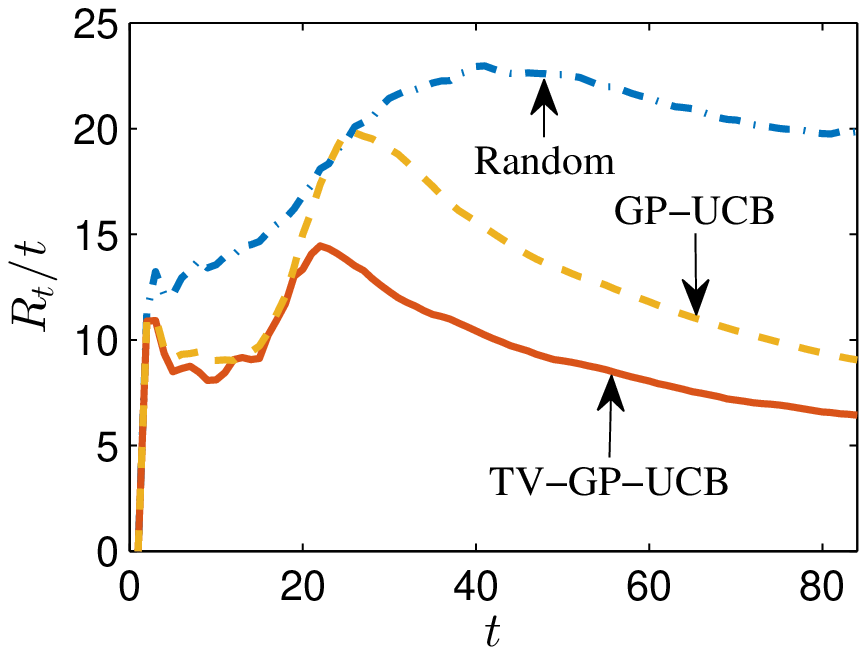}
	\caption{Traffic data, day 2}
	\label{fig:td_day2}  
  \end{subfigure}
\setcounter{subfigure}{4}
	\begin{subfigure}[b]{0.3\textwidth}
	 \includegraphics[width=\textwidth]{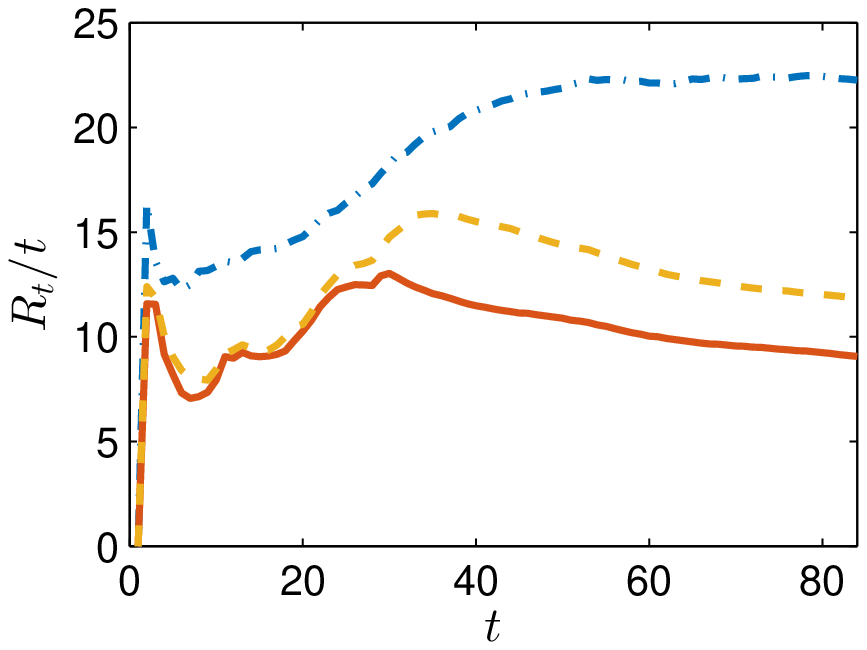}  
	\caption{Traffic data, day 4}
	\label{fig:td_day4}
  \end{subfigure}
\setcounter{subfigure}{1}
  \begin{subfigure}[b]{0.3\textwidth}
	\includegraphics[width=\textwidth]{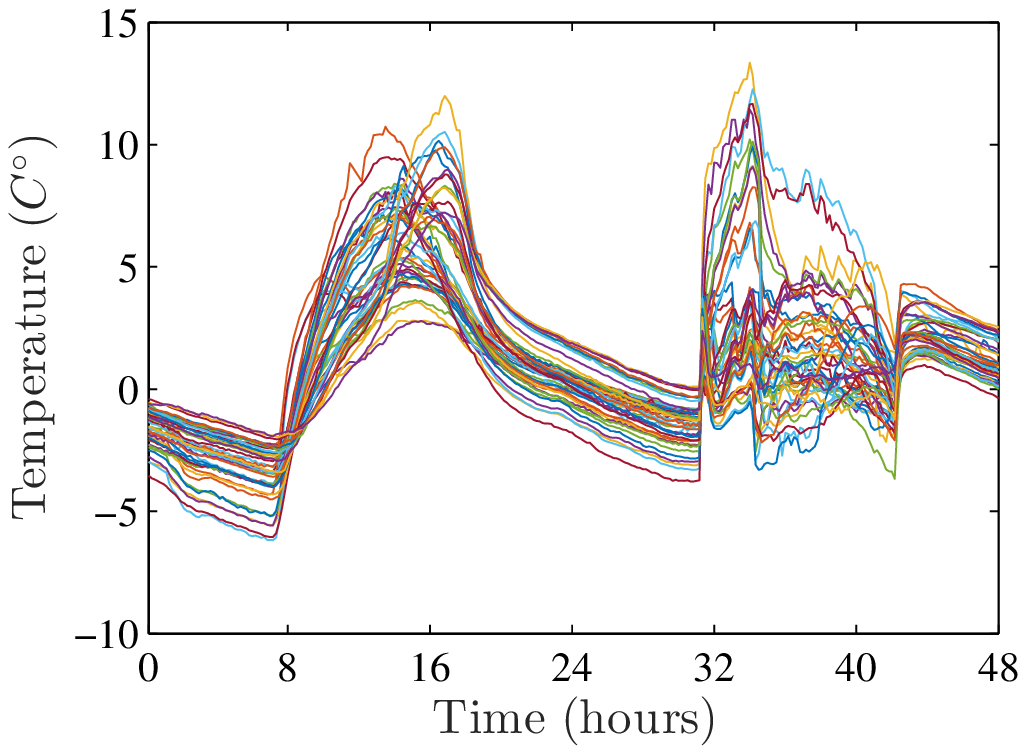}
	\caption{Temperature data}  
	\label{fig:test_data}	 
  \end{subfigure}
\setcounter{subfigure}{3}
  \begin{subfigure}[b]{0.3\textwidth}
   	\includegraphics[width=\textwidth]{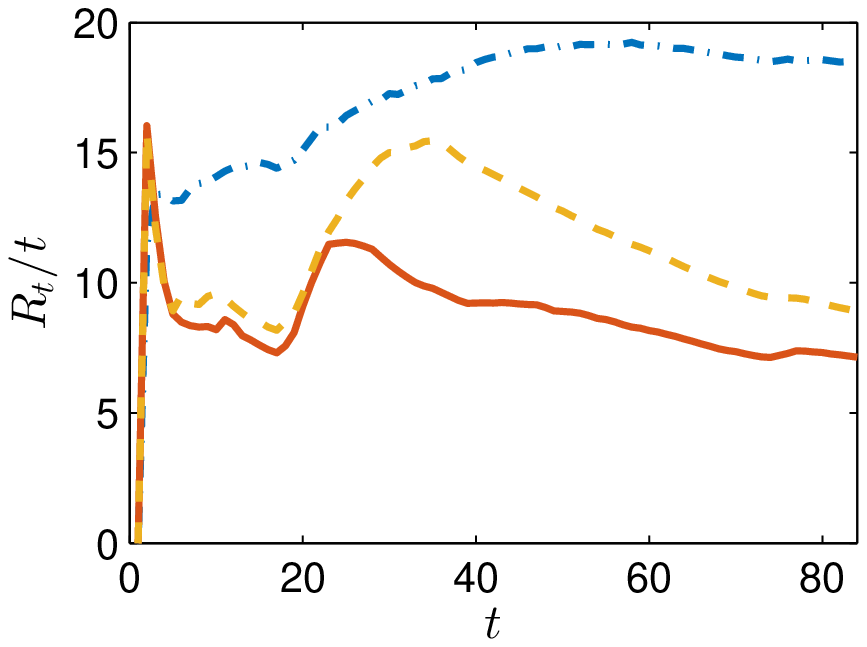}
	\caption{Traffic data, day 8}
	\label{fig:td_day8} 
  \end{subfigure} 
\setcounter{subfigure}{5}
  \begin{subfigure}[b]{0.3\textwidth}
	 \includegraphics[width=\textwidth]{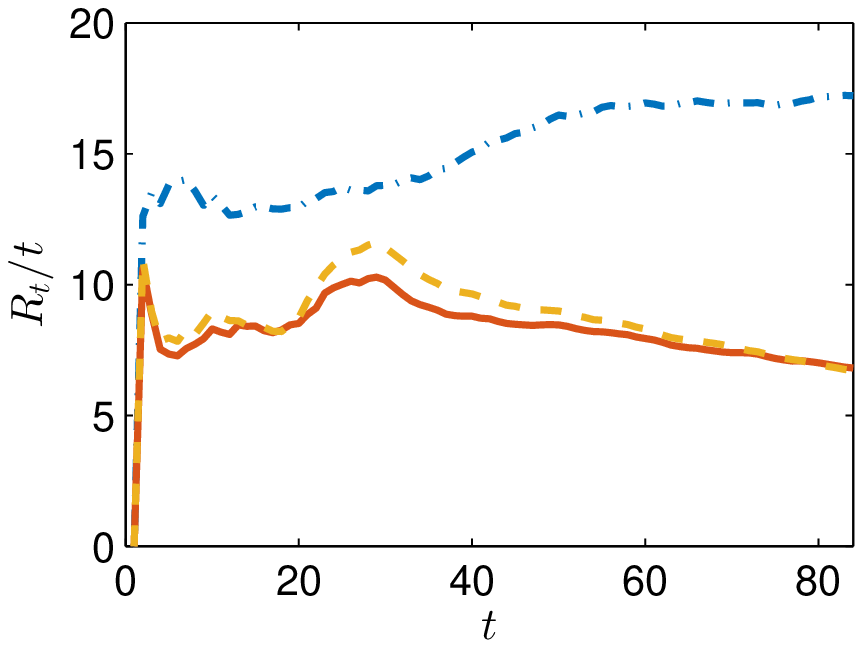}  
	\caption{Traffic data, day 10}
	\label{fig:td_day10}
  \end{subfigure} 
  \caption{Numerical performance of upper confidence bound algorithms on real data.}
  \label{fig:main}
\end{figure*}

\subsection{Real Data}
\label{r_s_data}

We use temperature data collected from $46$ sensors deployed at Intel Research, Berkeley. The dataset contains $5$ days of measurements collected at $10$-minute intervals. The goal of the spatiotemporal monitoring problem (see \cite{Kra11} for details) is to activate a sensor at every time step that reports a high temperature. Hence, $f_t$ consists of the set of all sensor temperature reportings at time $t$. A single sensor is activated every $10$ minutes, and the regret is measured as the temperature difference between reporting of the activated sensor and the one that reports the maximum temperature at that particular time.  Figure \ref{fig:test_data} plots each of the 46 functions with respect to time.

As a base comparison, we consider an algorithm that simply picks the sensors uniformly at random.  We also consider the standard GP-UCB algorithm \cite{Sri12}, even though it is unsuitable here since the reward function is varying with time.\footnote{In \cite{Sri12}, the same data was used to test GP-UCB in a different way; in each experiment, the function $f(x)$ was taken to be the set of temperatures at a single time.}  Although it is not shown, we note that the RExp3 algorithm \cite{Bes14} (Exp3 with resetting) performed comparably to GP-UCB for this data set, suffering from the fact that it does not exploit correlations between the sensors.

We use the first three days of measurements for learning our algorithms' parameters.  First, we  compute the empirical covariance matrix from these days and use it as the kernel matrix in all of the algorithms.  Next, using the same three training days, we obtain $\epsilon = 0.03$ by maximizing the marginal likelihood \cite{Van12}, and we obtain $N = 15$ by cross-validation.  The algorithms are run on the final two days of the data.
The results ($c_1 = 0.8, c_2 = 0.4, \sigma^2 = 0.5$ or $5\%$ of the signal variance) are shown in Figure \ref{fig:real_data}.  We observe that GP-UCB performs well for a short time, but then starts to suffer from the stale data, eventually becoming barely better than a random guess.  Once again, TV-GP-UCB improves over R-GP-UCB, with the gap generally increasing over the duration of the experiment.

Next, we use traffic speed data from $357$ sensors deployed along the highway I-880 South (California).  The dataset contains one month of measurements, where $84$ measurements were made on every day in between $6$ AM and $11$ AM.  Our goal is to identify the least congested part of the highway by tracking the point of maximum speed.  We use two thirds of the dataset to compute the empirical covariance matrix (and set it as the kernel matrix), and to learn $\epsilon$ by maximizing the marginal likelihood for all the training days \cite{Van12}, treating each day as being statistically independent. The last $10$ days were used for testing.  Due to the small time horizon $T = 84$ in comparison to the number of sensors, we restrict the domain to contain $50$ sensors, chosen randomly from the $357$. Our results ($\epsilon = 0.04$, $\sigma^2 = 5.0$ or $5\%$ of the signal variance, $T = 84$, $c_1 = 0.2$, $c_2 = 0.4$) were averaged over $20$ different initially activated sensors.   

In Figure~\ref{fig:main}, in final two columns, we show the outcome of the experiment for $4$ testing days (for the results on the rest of the days see Appendix~\ref{sec:TRAFFIC_RESULTS}). TV-GP-UCB outperforms GP-UCB on most testing days, with the two being comparable for a few of the days (e.g., see Figure \ref{fig:td_day10}).  The latter situation arises when the indices of the best sensors do not change drastically over the time horizon, which is not always the case.  In general, both algorithms suffer a large regret when sensors that were reporting high speeds suddenly change and start to report small speeds. However, TV-GP-UCB recovers more quickly from this compared to GP-UCB, due to its forgetting mechanism. 

Note that we have omitted R-GP-UCB from this experiment, since we found it to be unsuitable due to the small time horizon.  Moreover, this is the same reason that GP-UCB performs reasonably, unlike the temperature sensor example.  Essentially, GP-UCB suffers more with a longer time horizon due to the larger amount of stale data.

\section{Conclusion}

We have studied the bandit optimization problem with time-varying rewards, taking a new approach based on a GP that evolves according to a simple Markov model.  We introduced the R-GP-UCB and TV-GP-UCB algorithms, which, in contrast to previous algorithms, simultaneously trade off forgetting and remembering while also exploiting both spatial and temporal correlations.  Our regret bounds for these algorithms provide, to our knowledge, the first explicit characterizations of the trade-off between the time horizon $T$ and rate at which the function varies $\epsilon$ in a bandit setting. We also provided an algorithm-independent bound revealing that a linear dependence on $T$ for fixed $\epsilon$ is unavoidable.  Despite the simplicity of our theoretical model, we saw that the algorithms performed well on real world data sets that need not be matched to this model.


An immediate direction for future research is to determine to what extent the dependence on $\epsilon$ can be improved in our upper and lower bounds.  Moreover, one could move to the non-Bayesian setting and consider classes of time-varying functions whose smoothness is dictated by an RKHS norm; see \cite{Sri12} for the time-invariant counterpart.  Furthermore, while our time-varying model is primarily suited to handling steady changes, it could potentially be made even more effective by explicitly handling \emph{sudden} changes, e.g., by a combination of our techniques with those from previous works studying changepoint detection \cite{Ada07,Gar09}.

\section*{Acknowledgment}

We thank Andreas Krause for sharing his datasets with us for use in Section \ref{sec:NUMERICAL}.

This work was supported in part by the European Commission under Grant ERC Future Proof, SNF 200021-146750 and SNF CRSII2-147633, and `EPFL Fellows' Horizon2020 grant 665667.

\bibliographystyle{IEEEtran}
\bibliography{refs}

\begin{thebibliography}{10}
\providecommand{\url}[1]{#1}
\csname url@samestyle\endcsname
\providecommand{\newblock}{\relax}
\providecommand{\bibinfo}[2]{#2}
\providecommand{\BIBentrySTDinterwordspacing}{\spaceskip=0pt\relax}
\providecommand{\BIBentryALTinterwordstretchfactor}{4}
\providecommand{\BIBentryALTinterwordspacing}{\spaceskip=\fontdimen2\font plus
\BIBentryALTinterwordstretchfactor\fontdimen3\font minus
  \fontdimen4\font\relax}
\providecommand{\BIBforeignlanguage}[2]{{%
\expandafter\ifx\csname l@#1\endcsname\relax
\typeout{** WARNING: IEEEtran.bst: No hyphenation pattern has been}%
\typeout{** loaded for the language `#1'. Using the pattern for}%
\typeout{** the default language instead.}%
\else
\language=\csname l@#1\endcsname
\fi
#2}}
\providecommand{\BIBdecl}{\relax}
\BIBdecl

\bibitem{Sri12}
N.~Srinivas, A.~Krause, S.~Kakade, and M.~Seeger, ``Information-theoretic
  regret bounds for {G}aussian process optimization in the bandit setting,''
  \emph{IEEE Trans. Inf. Theory}, vol.~58, no.~5, pp. 3250--3265, May 2012.

\bibitem{Bub12}
S.~Bubeck and N.~Cesa-Bianchi, \emph{Regret Analysis of Stochastic and
  Nonstochastic Multi-Armed Bandit Problems}, ser. Found. Trend. Mach.
  Learn.\hskip 1em plus 0.5em minus 0.4em\relax Now Publishers, 2012.

\bibitem{Lai85}
T.~Lai and H.~Robbins, ``Asymptotically efficient adaptive allocation rules,''
  \emph{Adv. App. Math.}, vol.~6, no.~1, pp. 4 -- 22, 1985.

\bibitem{Bro10}
E.~Brochu, V.~M. Cora, and N.~de~Freitas, ``A tutorial on bayesian optimization
  of expensive cost functions, with application to active user modeling and
  hierarchical reinforcement learning,'' 2010, http://arxiv.org/abs/1012.2599.

\bibitem{Sno12}
J.~Snoek, H.~Larochelle, and R.~P. Adams, ``Practical {B}ayesian optimization
  of machine learning algorithms,'' in \emph{Adv. Neur. Inf. Proc. Sys.}, 2012.

\bibitem{Ras06}
C.~E. Rasmussen, ``Gaussian processes for machine learning.''\hskip 1em plus
  0.5em minus 0.4em\relax MIT Press, 2006.

\bibitem{Kra11}
A.~Krause and C.~S. Ong, ``Contextual {G}aussian process bandit optimization,''
  in \emph{Adv. Neur. Inf. Proc. Sys.}\hskip 1em plus 0.5em minus 0.4em\relax
  Curran Associates, Inc., 2011, pp. 2447--2455.

\bibitem{Djo13}
J.~Djolonga, A.~Krause, and V.~Cevher, ``High-dimensional {G}aussian process
  bandits,'' in \emph{Adv. Neur. Inf. Proc. Sys.}, C.~Burges, L.~Bottou,
  M.~Welling, Z.~Ghahramani, and K.~Weinberger, Eds.\hskip 1em plus 0.5em minus
  0.4em\relax Curran Associates, Inc., 2013, pp. 1025--1033.

\bibitem{Sno14}
J.~Snoek, K.~Swersky, R.~Zemel, and R.~P. Adams, ``Input warping for {B}ayesian
  optimization of non-stationary functions,'' in \emph{Int. Conf. Mach.
  Learn.}, 2014.

\bibitem{Wan13}
Z.~Wang, M.~Zoghi, F.~Hutter, D.~Matheson, and N.~de~Freitas, ``Bayesian
  optimization in high dimensions via random embeddings,'' in \emph{Int. Joint.
  Conf. Art. Int.}, 2013.

\bibitem{Bub15}
S.~Bubeck, O.~Dekel, T.~Koren, and Y.~Peres, ``Bandit convex optimization:
  $\sqrt{T}$ regret in one dimension,'' 2015, http://arxiv.org/abs/1502.06398.

\bibitem{Whi88}
P.~Whittle, ``\BIBforeignlanguage{English}{Restless bandits: Activity
  allocation in a changing world},'' \emph{\BIBforeignlanguage{English}{J. App.
  Prob.}}, vol.~25, pp. 287--298, 1988.

\bibitem{Ber00}
D.~Bertsimas and J.~Niño-Mora, ``Restless bandits, linear programming
  relaxations, and a primal-dual index heuristic,'' \emph{Operations Research},
  vol.~48, no.~1, pp. 80--90, 2000.

\bibitem{Ort12}
R.~Ortner, D.~Ryabko, P.~Auer, and R.~Munos,
  ``\BIBforeignlanguage{English}{Regret bounds for restless {M}arkov
  bandits},'' in \emph{\BIBforeignlanguage{English}{Algorithmic Learning
  Theory}}.\hskip 1em plus 0.5em minus 0.4em\relax Springer Berlin Heidelberg,
  2012, pp. 214--228.

\bibitem{Sli08}
A.~Slivkins and E.~Upfal, ``Adapting to a changing environment: the {B}rownian
  restless bandits,'' in \emph{Conf. Learn. Theory}, 2008.

\bibitem{Tek12}
C.~Tekin and M.~Liu, ``Online learning of rested and restless bandits,''
  \emph{IEEE Trans. Inf. Theory}, vol.~58, no.~8, pp. 5588--5611, Aug. 2012.

\bibitem{Liu13}
H.~Liu, K.~Liu, and Q.~Zhao, ``Learning in a changing world: Restless
  multiarmed bandit with unknown dynamics,'' \emph{IEEE Trans. Inf. Theory},
  vol.~59, no.~3, pp. 1902--1916, March 2013.

\bibitem{Bes14}
O.~Besbes, Y.~Gur, and A.~Zeevi, ``Stochastic multi-armed-bandit problem with
  non-stationary rewards,'' in \emph{Adv. Neur. Inf. Proc. Sys.}, 2014, pp.
  199--207.

\bibitem{Van12}
S.~Van~Vaerenbergh, M.~L{\'a}zaro-Gredilla, and I.~Santamar{\'\i}a, ``Kernel
  recursive least-squares tracker for time-varying regression,'' \emph{IEEE
  Trans. Neur. Net. Learn. Sys.}, vol.~23, no.~8, pp. 1313--1326, 2012.

\bibitem{Van12a}
S.~Van~Vaerenbergh, I.~Santamar{\'\i}a, and M.~L{\'a}zaro-Gredilla,
  ``Estimation of the forgetting factor in kernel recursive least squares,'' in
  \emph{IEEE. Int. Workshop Mach. Learn. SIg. Proc.}, 2012, pp. 1--6.

\bibitem{Osb08}
M.~A. Osborne, S.~Roberts, A.~Rogers, S.~Ramchurn, and N.~R. Jennings,
  ``Towards real-time information processing of sensor network data using
  computationally efficient multi-output gaussian processes,'' in \emph{Proc.
  Int. Conf. Inf. Proc. Sens. Net.}, 2008, pp. 109--120.

\bibitem{Kan15}
K.~Kandasamy, J.~G. Schneider, and B.~P{\'{o}}czos, ``High dimensional
  {B}ayesian optimisation and {B}andits via additive models,'' 2015,
  http://arxiv.org/abs/1503.01673.

\bibitem{Ada07}
R.~P. Adams and D.~J. MacKay, ``Bayesian online changepoint detection,'' 2007,
  http://arxiv.org/abs/0710.3742.

\bibitem{Gar09}
R.~Garnett, M.~A. Osborne, and S.~J. Roberts, ``Sequential bayesian prediction
  in the presence of changepoints,'' in \emph{Proc. Inf. Conf. Mach. Learn.},
  2009.

\bibitem{Cov01}
T.~M. Cover and J.~A. Thomas, \emph{Elements of Information Theory}.\hskip 1em
  plus 0.5em minus 0.4em\relax John Wiley \& Sons, Inc., 2001.

\bibitem{Hor85}
R.~A. Horn and C.~R. Johnson, \emph{Matrix Analysis}, 2nd~ed.\hskip 1em plus
  0.5em minus 0.4em\relax New York, NY, USA: Cambridge University Press, 2012.

\bibitem{Bha97}
R.~Bhatia, \emph{Matrix Analysis}.\hskip 1em plus 0.5em minus 0.4em\relax
  Springer, 1997.

\end{thebibliography}

\newpage
\onecolumn

\appendix

\begin{centering}
    {\Large \bf Supplementary Material for ``Time-Varying Gaussian Process Bandit Optimization''}
    
    {\large \bf (AISTATS 2016; Ilija Bogunovic, Volkan Cevher, Jonathan Scarlett)} \par
\end{centering}

Note that all citations here are to the bibliography in the main document, and similarly for many of the cross-references.

\section{Posterior Updates} \label{sec:TV_UPDATES_DERIV}

Here we derive the posterior update rules for the time-varying setting via a suitable adaptation of the derivation for the time-invariant setting \cite{Ras06}.  Observe from \eqref{eq:tv_model0}--\eqref{eq:tv_model} that each function $f_t$ depends only on the functions $g_i$ for $i \le t$.  By a simple recursion, we readily obtain for all $t$, $j$ and $x$ that
\begin{equation}
    \cov[f_t(x)f_{t+j}(x')] = (1-\epsilon)^{j/2} \myexpect[f_t(x)f_t(x')] =  (1-\epsilon)^{j/2} k(x,x').
\end{equation}
Hence, and since each output $y_t$ equals the corresponding function sample $f_t(x_t)$ plus additive Gaussian noise $z_t$ with variance $\sigma^2$, the joint distribution between the previous outputs $\vy_t=(y_1,\dotsc,y_t)$ (corresponding to the points $\vx_t = (x_1,\dotsc,x_t)$) and the next function value $f_{t+1}(x)$ is 
\begin{equation}
\begin{bmatrix}
       \vy           \\[0.5em]
       f_{t+1}(x)
     \end{bmatrix}
     \sim \mathcal{N} \left(\mathbf{0}, 
     \begin{bmatrix}
      \vKtil_t + \sigma^2\vI_t & \vktil_t(x)            \\[0.5em]
       \vktil_t(x)^T & k(x,x)
     \end{bmatrix}
      \right)
\end{equation}
using the definitions in the proposition statement.  Using the formula for the conditional distribution associated with a jointly Gaussian random vector \cite[App.~A]{Ras06}, we find that $f_{t+1}(x)$ is conditionally Gaussian with mean $\mutil_t(x)$ and variance $\sigmatil_t(x)^2$, as was to be shown. 

\section{Learning $\epsilon$ via Maximum-Likelihood} \label{sec:OPT_EPS}

In this section, we provide an overview of how $\epsilon$ can be learned from training data in a principled manner; the details can be found in \cite[Section~4.3]{Van12a} and \cite[Section~5]{Ras06}.	 Throughout this appendix, we assume that the kernel matrix is parametrized by a set of hyperparameters $\theta$ (e.g., $\theta = (\nu,l)$ for the M\'atern kernel), $\sigma$ and $\epsilon$.

Let $\bar{\vy}$ be a vector of observations such that the $i$-th entry is observed at time $t_i$ as a result of sampling the function $f_{t_i}$ at location $x_i$.  Note that there will typically be many indices $i$ sharing common values of $t_i$, since in the training data we often have multiple samples at each time.  Under our time-varying GP model, the marginal log-likelihood of $\bar{\vy}$ given the hyperparameters is
\begin{equation}
	\log p(\bar{\vy} | \theta, \sigma, \epsilon) =  - \dfrac{1}{2} \bar{\vy}^T (\bar{\vK} \circ \bar{\vD} + \sigma^2 \vI)^{-1} \bar{\vy} - \dfrac{1}{2} \log |\bar{\vK} \circ \bar{\vD} + \sigma^2 \vI | - \dfrac{n}{2} \log(2\pi)
\end{equation}
where $(\bar{\vK})_{ij} = k(x_i,x_j)$ and $(\bar{\vD})_{ij} = (1-\epsilon)^{|t_i - t_j|/2}$.  To set the hyperparameters by maximizing the marginal likelihood, we can use the partial derivatives with respect to the hyperparameters. In particular, we have
\begin{equation}
	\dfrac{\partial \log p(\bar{\vy} | \theta, \sigma, \epsilon) }{\partial \epsilon} = \dfrac{1}{2} \tr \big( (\balpha \balpha^T - (\bar{\vK} \circ \bar{\vD} + \sigma^2 \vI)^{-1}) (\bar{\vK} \circ \bar{\vD'}) \big),
\end{equation}
where $\balpha = (\bar{\vK} \circ \bar{\vD} + \sigma^2 \vI)^{-1} \bar{\vy}$, and  $(\bar{\vD}')_{ij} = -v (1- \epsilon)^{v - 1}$ with $v = |t_i - t_j|/2$.

We can now fit $\epsilon$ and the other hyperparameters by optimizing the marginal likelihood on the training data, e.g.  by using an optimization algorithm from the family of quasi-Newton methods.

\section{Analysis of TV-GP-UCB (Theorem \ref{thm:general})}

We recall the following alternative form for the mutual information (see \eqref{eq:MI}) from \cite[Lemma~5.3]{Sri12}, which extends immediately to the time-varying setting:
\begin{equation}
    \Itil(\vf_T;\vy_T) = \frac{1}{2}\sum_{t=1}^T \log\big(1 + \sigma^{-2} \sigmatil_{t-1}^2(x_t)\big). \label{eq:MI2}
\end{equation}

\subsection{Proof of \eqref{eq:general_bound}}

The initial steps of the proof follow similar ideas to \cite{Sri12}, but with suitable modifications to handle the fact that we have a different function $f_t$ at each time instant.  A key difficulty is in subsequently bounding the maximum mutual information in the presence of time variations, which is done in the following subsection.

We first fix a discretization $D_t \subset  D \subseteq [0,r]^d$ of size $(\tau_t)^d$ satisfying
\begin{equation}
	\| x - [x]_t \|_1 \leq rd / \tau_t, \quad \forall x \in D, \label{eq:gb1}
\end{equation}
where $[x]_t$ denotes the closest point in $D_t$ to $x$.  For example, a uniformly-spaced grid suffices to ensure that this holds.

We now fix a constant $\delta > 0$ and an increasing sequence of positive constants $\{\pi_t\}_{t=1}^{\infty}$ satisfying $\sum_{t \geq 1} \pi_t^{-1} = 1$ (e.g. $\pi_t = \pi^2 t^2 / 6$), and condition on three high-probability events:
\begin{enumerate}[leftmargin=5ex]
    \item We first claim that if $\beta_t \ge 2\log\frac{3\pi_t}{\delta}$ then the selected points $\{x_t\}_{t=1}^T$ satisfy the confidence bounds
    \begin{equation}
        | f_t(x_t) - \mutil_{t-1} (x_t) |  \leq \beta_t^{1/2} \sigmatil_{t-1} (x_t), \quad \forall t \label{eq:TV_assump1}
    \end{equation}
    with probability at least $1-\frac{\delta}{3}$.  To see this, we note that conditioned on the outputs $(y_1,\dotsc,y_{t-1})$, the sampled points $(x_1,\dotsc,x_t)$ are deterministic, and $f_t(x_t) \sim \Ndist(\mutil_{t-1}(x_t), \sigmatil_{t-1}^2(x_t))$.  An $\Ndist(\mu,\sigma^2)$ random variable is within $\sqrt{\beta}\sigma$ of $\mu$ with probability at least $1-e^{-\beta/2}$, and hence our choice of $\beta_t$ ensures that the event corresponding to time $t$ in \eqref{eq:TV_assump1} occurs with probability at least $\frac{\delta}{3\pi_t}$.  Taking the union bound over $t$ establishes the claim.
    \item By the same reasoning with an additional union bound over $x \in D_t$, if $\beta_t \ge 2\log\frac{3 |D_t| \pi_t}{\delta}$, then
    \begin{equation}
        | f_t(x) - \mutil_{t-1} (x) |  \leq \beta_t^{1/2} \sigmatil_{t-1}(x), \quad \forall t, x \in D_t \label{eq:TV_assump2}
    \end{equation}
    with probability at least $1-\frac{\delta}{3}$.  
    \item Finally, we claim that setting $L_t = b \sqrt{\log(3da \pi_t / \delta)}$ yields
    \begin{equation}
        | f_t(x) - f_t(x') | \leq L_t \|x - x' \|_1, \quad \forall t, x\in D, x' \in D \label{eq:TV_assump3}
    \end{equation}
    with probability at least $1-\frac{\delta}{3}$.  To see this, we note that by the assumption in \eqref{eq:k_deriv1} and the union bound over $j = 1,\dotsc,d$, the event corresponding to time $t$ in \eqref{eq:TV_assump3} holds with probability at least $1 - dae^{-L_t^2 / b^2} = \frac{\delta}{3\pi_t}$.  Taking the union bound over $t$ establishes the claim.
\end{enumerate} 

Again applying the union bound, all three of \eqref{eq:TV_assump1}--\eqref{eq:TV_assump3} hold with probability at least $1-\delta$.  We henceforth condition on each of them occurring.

Combining \eqref{eq:gb1} with \eqref{eq:TV_assump3} yields for all $x$ that
\begin{align}
    | f_t(x) - f_t([x]_t) | 
        &\le L_t rd / \tau_t \\
        &= b \sqrt{\log(3da \pi_t / \delta)} rd / \tau_t, 
\end{align}
and hence choosing $\tau_t = rdbt^2 \sqrt{\log(2da \pi_t / \delta)}$ yields
\begin{equation}
    | f_t(x) - f_t([x]_t) | \le 1/t^2. \label{eq:quant_diff}
\end{equation}
Note that this choice of $\tau_t$ yields $|D_t| = (\tau_t)^d = (rdbt^2\sqrt{\log(3da \pi_t / \delta)})^d$. In order to satisfy both lower bounds on $\beta_t$ stated before \eqref{eq:TV_assump1} and \eqref{eq:TV_assump2}, it suffices to take higher of the two (i.e.,~the second), yielding
\begin{equation}
	\beta_t = 2 \log (3 \pi_t / \delta) + 2d \log(rdbt^2\sqrt{\log(3da \pi_t / \delta)}).
\end{equation}
This coincides with \eqref{eq:beta_T} upon setting $\pi_t = \pi^2t^2 / 6$.

Substituting \eqref{eq:quant_diff} into \eqref{eq:TV_assump2} and applying the triangle inequality, we find the maximizing point $x_t^*$ at time $t$ satisfies
\begin{equation}
    |f_t(x_t^*) - \mutil_{t-1}([x_t^*]_t)| \leq \beta_t^{1/2} \sigmatil_{t-1}([x_t^*]_t) + 1/t^2. \label{eq:gb4}
\end{equation}
Thus, we can bound the instantaneous regret as follows:
\begin{align}
r_t & = f_t (x_t^*) - f_t(x_t) \\
	& \leq \mutil_{t-1}([x_t^*]_t) + \beta_t^{1/2} \sigmatil_{t-1}([x^*_t]_t) + 1/t^2 - f_t(x_t) \label{eq:gb5_1}\\
	& \leq \mutil_{t-1}(x_t) + \beta_t^{1/2} \sigmatil_{t-1}(x_t) + 1/t^2 - f_t(x_t)\label{eq:gb5_2}\\
	& \leq 2 \beta_t^{1/2} \sigmatil_{t-1}(x_t) + 1/t^2 \label{eq:gb5_3},
\end{align}
where in \eqref{eq:gb5_1} we used \eqref{eq:gb4}, \eqref{eq:gb5_2} follows since the function $\mutil_{t-1}(x) + \beta_t^{1/2}\sigmatil_{t-1}(x)$ is maximized at $x_t$ by the definition of the algorithm, and \eqref{eq:gb5_3} follows from \eqref{eq:TV_assump1}.

Finally, we bound the cumulative regret as
\begin{align}
	R_T = \sum_{t=1}^T r_t 
    	&\le \sum_{t=1}^T  \Big(2\beta_t^{1/2} \sigmatil_{t-1}(x_t) + 1/t^2\Big) \label{eq:gb6_1} \\
        &\le  \sqrt{T \sum_{t=1}^T  4 \beta_t \sigmatil^2_{t-1}(x_t)} + 2\label{eq:gb6_2} \\
	& \leq \sqrt{C_1 T \beta_T \gammatil_T} + 2 \label{eq:gb6_3},
\end{align}
where \eqref{eq:gb6_2} follows using $\sum_{t=1}^{\infty} 1/t^2 = \pi^2/6 \le 2$   the fact that $\|z\|_1 \le \sqrt{T}\|z\|_2$ for any vector $z \in \real^T$. Equation \eqref{eq:gb6_3} is proved using following steps from \cite[Lemma~5.4]{Sri12}, which we include for completeness:
\begin{align}
    \sum_{t=1}^T 4 \beta_t \sigmatil^2_{t-1}(x_t) 
        & \leq  4\beta_T \sigma^2 \sum_{t=1}^T \sigma^{-2} \sigmatil^2_{t-1}(x_t) \label{eq:C1_bound1} \\
        & \leq  4\beta_T \sigma^2 \sum_{t=1}^T  C_2 \log\big(1 + \sigma^{-2} \sigmatil^2_{t-1}(x_t)\big) \label{eq:C1_bound2} \\
        & \leq C_1 \beta_T \gammatil_T, \label{eq:C1_bound3}
\end{align} 
where \eqref{eq:C1_bound1} follows since $\beta_t$ is increasing in $T$, \eqref{eq:C1_bound2} holds with $C_2 = \sigma^{-2}/\log(1+\sigma^{-2})$ using the identity $z^2 \le C_2 \log(1+z^2)$ for $z^2\in[0,\sigma^{-2}]$ (note also that $\sigma^{-2}\sigmatil^2_{t-1}(\vx_t) \le \sigma^{-2}k(\vx_t,\vx_t) \le \sigma^{-2}$), and \eqref{eq:C1_bound3} follows from the definitions of $C_1$ and $\gammatil_T$, along with the alternative form for the mutual information in \eqref{eq:MI2}.

\subsection{Proof of \eqref{eq:weakened2}}

It remains to show that
\begin{equation}
    \gammatil_T \le \bigg(\frac{T}{\Ntil} + 1\bigg) \Big(\gammaNtilTI + \Ntil^3 \epsilon\Big)
\end{equation} 
under the definitions in \eqref{eq:MI}--\eqref{eq:gamma}.  Recall that $\vx_T=(x_1,\dotsc,x_T)$ are the points of interest, $\vf_T = (f_1(x_1),\dotsc,f_T(x_T))$ are the corresponding function values, and $\vy_T = (y_1,\dotsc,y_T)$ contains the corresponding noisy observations with $y_i = f_i(x_i) + z_i$.

At a high level, we bound the mutual information with time variations in terms of the corresponding quantity for the time-invariant case \cite{Sri12} by splitting the time steps $\{1,\dotsc,T\}$ into $\frac{T}{\Ntil}$ blocks of length $\Ntil$, such that within each block the function $f_i$ does not vary significantly.  We assume for the time being that $T/\Ntil$ is an integer, and then handle the general case.

Using the chain rule for mutual information and the fact that the noise sequence $\{z_i\}$ is independent, we have \cite[Lemma 7.9.2]{Cov01} 
\begin{equation}
	\Itil(\vf_T;\vy_T) \le \sum_{i=1}^{T/\Ntil} \Itil(\vf_{\Ntil}^{(i)};\vy_{\Ntil}^{(i)}),
\end{equation}
where $\vy_{\Ntil}^{(i)} = (y_{\Ntil(i-1)+1},\dotsc,y_{\Ntil i})$ contains the measurements in the $i$-th block, and $\vf_{\Ntil}^{(i)}$ is defined analogously.  Maximizing both sides over $(x_1,\dotsc,x_T$, we obtain
\begin{equation}
	\gammatil_T \le \frac{T}{\Ntil} \gammatil_{\Ntil}. \label{eq:gamma_split}
\end{equation}
We are left to bound $\gammatil_{\Ntil}$.  To this end, we write the relevant covariance matrix as
\begin{equation}
	\vKtil_{\Ntil} = \vK_{\Ntil} \circ \vD_{\Ntil} = \vK_{\Ntil} + \vA_{\Ntil},
\end{equation}
where
\begin{align}
	\vA_{\Ntil} &:= \vK_{\Ntil} \circ \vD_{\Ntil} - \vK_{\Ntil} \\
          &= \vK_{\Ntil} \circ (\vD_{\Ntil} - \vone_{\Ntil}) \label{eq:AN_def2}
\end{align}
and $\vone_{\Ntil}$ is the $\Ntil \times \Ntil$ matrix of ones.  Observe that the $(i,j)$-th entry of $\vD_{\Ntil} - \vone_{\Ntil}$ has absolute value $1-(1-\epsilon)^{\frac{|i-j|}{2}}$, which is upper bounded for all $\epsilon\in[0,1]$ by $\epsilon|i-j|$.\footnote{For $|i-j| \ge 2$, this follows since the function of interest is concave, passes through the origin, and has derivative $\frac{|i-j|}{2} \le |i-j|$ there.  For $k=1$, the statement follows by observing that equality holds for $\epsilon\in\{0,1\}$, and noting that the function of interest is convex.}  Hence, and using the fact that each entry of $\vK_{\Ntil}$ lies in the range $[0,1]$, we obtain the following bound on the Frobenius norm:
\begin{align}
    \|\vA_{\Ntil}\|_F^2 &\le \sum_{i,j} (i-j)^2 \epsilon^2 \\
                          &= \frac{1}{6}\Ntil^2(\Ntil^2-1)\epsilon^2 \label{eq:double_sum} \\
                          &\le \Ntil^4\epsilon^2, \label{eq:frob_bound}
\end{align}
where \eqref{eq:double_sum} is a standard double summation formula. We will use this inequality to bound $\gammatil_N$ via Mirsky's theorem, which is given as follows.
\begin{lemma} \emph{(Mirsky's theorem \cite[Cor.~7.4.9.3]{Hor85})}
    For any matrices $\vU_{\Ntil}$ and $\vV_{\Ntil}$, and any unitarily invariant norm $\vertiii{\cdot}$, we have 
    \begin{equation} 
        \vertiii{\mathrm{diag}(\lambda_1(\vU_{\Ntil}),\dotsc,\lambda_{\Ntil}(\vU_{\Ntil})) - \mathrm{diag}(\lambda_1(\vV_{\Ntil}),\dotsc,\lambda_{\Ntil}(\vV_{\Ntil}))} \le \vertiii{\vU_{\Ntil} - \vV_{\Ntil}}, 
    \end{equation}
    where $\lambda_i(\cdot)$ is the $i$-th largest eigenvalue.
\end{lemma}
Using this lemma with $\vU_{\Ntil} = \vK_{\Ntil} + \vA_{\Ntil}$, $\vV_{\Ntil} = \vK_{\Ntil}$, and $\vertiii{\cdot} = \|\cdot\|_F$, and making use of \eqref{eq:frob_bound}, we find that $\lambda_i(\vK_{\Ntil} + \vA_{\Ntil}) = \lambda_i(\vK_{\Ntil}) + \Delta_i$ for some $\{\Delta_i\}_{i=1}^{\Ntil}$ satisfying $\sum_{i=1}^{\Ntil} \Delta_i^2 \le \Ntil^4 \epsilon^2$.  We thus have
\begin{align}
    \gammatil_{\Ntil} &= \sum_{i=1}^{\Ntil} \log\big( 1 + \lambda_i(\vK_{\Ntil} + \vA_{\Ntil}) \big) \\
             &= \sum_{i=1}^{\Ntil} \log\big( 1 + \lambda_i(\vK_{\Ntil}) + \Delta_i \big) \\
             &\le \gammaNtilTI + \sum_{i=1}^{\Ntil} \log(1+\Delta_i) \label{eq:gamma_s3} \\
             &\le \gammaNtilTI + \Ntil\log(1+\Ntil^2\epsilon) \label{eq:gamma_s4} \\
             &\le \gammaNtilTI + \Ntil^3\epsilon, \label{eq:gamma_s5}
\end{align}
where \eqref{eq:gamma_s3} follows from the inequality $\log(1+a+b) \le \log(1+a) + \log(1+b)$ for non-negative $a$ and $b$ (and the definition in \eqref{eq:MI_TI}), \eqref{eq:gamma_s4} follows since a simple analysis of the optimality conditions of 
\begin{equation}
    \mathrm{maximize} \quad \sum_{i=1}^{\Ntil} \log(1+\Delta_i) \qquad \mathrm{subject~to} \qquad \sum_{i=1}^{\Ntil} \Delta_i^2 \le \Ntil^4 \epsilon^2
\end{equation}
reveals that the maximum is achieved when all of the $\Delta_i$ are equal to $\Ntil^2\epsilon$, and \eqref{eq:gamma_s5} follows from the inequality $\log(1+a) \le a$.

Recalling that we are considering the case that $T/\Ntil$ is an integer, we obtain \eqref{eq:weakened2} by combining \eqref{eq:gamma_split} and \eqref{eq:gamma_s5}.  In the general case, we simply use the fact that $\gammatil_T$ is increasing in $T$ by definition, hence leading to the addition of one in \eqref{eq:weakened2}.

\section{Analysis of R-GP-UCB (Theorem \ref{thm:general_R})}

Parts of the proof of Theorem \ref{thm:general_R} overlap with that of Theorem \ref{thm:general}; we focus primarily on the key differences.  First, overloading the notation from the TV-GP-UCB analysis, we let $\mutil_t(x)$ and $\sigmatil_t(x)$ be defined as in \eqref{eq:mu_update}--\eqref{eq:sigma_update}, but using \emph{only the samples since the previous reset in the R-GP-UCB algorithm}, and similarly for $\vk_t$, $\vktil_t$, $\vd_t$, and so on.  Thus, for example, the dimension of $\vk_t$ is at most the length $N$ between resets, and the entries of $\vD_t$ are no smaller than $(1-\epsilon)^{N/2}$.  Note that the time-invariant counterparts $\muTI_t(x)$ and $\sigmaTI_t(x)$ (computed using $\vk$ and $\vK$ in place of $\vktil$ and $\vKtil$) are used in the algorithm, thus creating a mismatch that must be properly handled.

Recall the definitions of the discretization $D_t$ (whose cardinality is again set to $\tau_t^d$ for some $\tau_t$), the corresponding quantization function $[x]_t$, and the constants $\pi_t$. We now condition on four (rather than three) high probability events:
\begin{itemize}[leftmargin=5ex]
    \item Setting $\beta_t = 2\log\frac{4|D_t| \pi_t }{\delta}$, the same arguments as those leading to \eqref{eq:TV_assump1}--\eqref{eq:TV_assump2} reveal that
        \begin{gather}
           |f_t(x_t) - \mutil_{t-1}(x_t)| \le \beta_t^{1/2} \sigmatil_{t-1}(x_t) \quad \forall t \ge 1 \label{eq:R_assump0} \\
           |f_t(x) - \mutil_{t-1}(x)| \le \beta_t^{1/2} \sigmatil_{t-1}(x) \quad \forall t \ge 1, x \in D_t \label{eq:R_assump1}
       \end{gather}
    with probability at least $1 - \frac{\delta}{2}$.  Note that in proving these claims we only condition on the observations since the last reset, rather than all of the points since $t=1$.
    \item Using the same argument as \eqref{eq:TV_assump3}, the assumption in \eqref{eq:k_deriv1} implies that
    \begin{equation}
        \Big| \frac{\partial f_t(x)}{\partial x^{(j)}} \Big| \le L_t := b_1 \sqrt{\log \frac{4da_1\pi_t}{\delta}} \quad \forall t \ge 1, x \in D, j\in\{1,\dotsc,d\} \label{eq:R_assump2}
    \end{equation}
    with probability at least  $1 - \frac{\delta}{4}$.
    \item We claim that the assumption in \eqref{eq:k_deriv0} similarly implies that
    \begin{equation}
         \big| y_t \big| \le \Ltil_t := (2+b_0) \sqrt{\log \frac{4(1+a_0)\pi_t}{\delta}} \quad \forall t \ge 1, x \in D \label{eq:R_assump3}
    \end{equation}
    with probability at least  $1 - \frac{\delta}{4}$.  To see this, we first note that $\Pr\big( |z_t| \le L \big) \le e^{-L^2/2}$ since $z_t \sim \Ndist(0,1)$, and by a standard bound on the standard normal tail probability.  Combining this with \eqref{eq:k_deriv0} and noting that $|y_t| \le |f_t(x_t)| + |z_t|$, we find that $\Pr\big( |y_t| > 2L \big)$ is upper bounded by $e^{-L^2/2} + a_0 e^{-(L/b_0)^2}$, which in turn is upper bounded by $(1+a_0) e^{-(L/(2+b_0))^2}$.  Choosing $L = \Ltil_t/2$ and applying the union bound and some simple manipulations, we obtain \eqref{eq:R_assump3}.
\end{itemize}

By the union bound, all four of \eqref{eq:R_assump0}--\eqref{eq:R_assump3} hold with probability at least $1-\delta$.

As in the TV-GP-UCB proof, we set $\tau_t = rdt^2 L_t$, thus ensuring that $|f_t(x) - f_t([x]_t)| \le \frac{1}{t^2}$ for all $x \in D$.  Defining
\begin{align}
    \Delta_{t}^{(\mu)} := \sup_{x\in D} |\mutil_t(x) - \muTI_t(x)| \label{eq:Delta_mu} \\
    \Delta_{t}^{(\sigma)} := \sup_{x\in D} |\sigmatil_t(x) - \sigmaTI_t(x)|  \label{eq:Delta_sigma}
\end{align}
to be the maximal errors between the true and the mismatched posterior updates, we have the following:
\begin{align}
    r_t &= f_t(x_t^*) - f_t(x_t)  \\
         &\le f_t([x^*_t]_t) - f_t(x_t) + \frac{1}{t^2} \label{eq:R_rt_2} \\
         &\le \mutil_{t-1}([x^*_t]_t) + \beta_t^{1/2} \sigmatil_{t-1}([x_t^*]_t) - \mutil_{t-1}(x_t) + \beta_t^{1/2} \sigmatil_{t-1}(x_t) + \frac{1}{t^2} \label{eq:R_rt_3} \\
         &\le \muTI_{t-1}([x^*_t]_t) + \beta_t^{1/2} \sigmaTI_{t-1}([x_t^*]_t) - \muTI_{t-1}(x_t) + \beta_t^{1/2} \sigmaTI_{t-1}(x_t) + 2\Delta_{t}^{(\mu)} + 2\beta_t^{1/2}\Delta_{t}^{(\sigma)} + \frac{1}{t^2} \label{eq:R_rt_4} \\
         &\le 2\beta_t^{1/2} \sigmaTI_{t-1}(x_t) + 2\Delta_{t}^{(\mu)} + 2\beta_t^{1/2}\Delta_{t}^{(\sigma)} + \frac{1}{t^2}, \label{eq:R_rt_5}
\end{align}
where \eqref{eq:R_rt_2} follows in the same way as \eqref{eq:gb5_1}, \eqref{eq:R_rt_3} follows from \eqref{eq:R_assump0}--\eqref{eq:R_assump1}, \eqref{eq:R_rt_4} follows from the definitions in \eqref{eq:Delta_mu}--\eqref{eq:Delta_sigma}, and \eqref{eq:R_rt_5} follows from the choice of $x_t$ in the algorithm.

The key remaining step is to characterize $\Delta_{t}^{(\mu)}$ and $\Delta_{t}^{(\sigma)}$.  Our findings are summarized in the following lemma.

\begin{lemma} \label{lem:Deltas}
    Conditioned on the event in \eqref{eq:R_assump3}, we have $\Delta_{t}^{(\mu)} \le \big(\sigma^{-2} + \sigma^{-4}\big) N^3 \epsilon \Ltil_t$ and  $\Delta_{t}^{(\sigma)} \le \big(3\sigma^{-2} + \sigma^{-4}\big) N^3 \epsilon$ almost surely.
\end{lemma}

This lemma implies Theorem \ref{thm:general_R} upon substitution into \eqref{eq:R_rt_5}, setting $\pi_t = \pi^2 t^2 / 6$, and following the steps from \eqref{eq:gb6_1} onwards.  In the remainder of the section, we prove the lemma.  The claims on $\Delta_{t}^{(\mu)}$ and $\Delta_{t}^{(\sigma)}$ are proved similarly; we focus primarily on the latter since it is the (slightly) more difficult of the two.

The subsequent analysis applies for arbitrary values of $t$ and $x$, so we use the shorthands $\vk := \vk_t(x)$, $\vK := \vK_t(x)$,  $\vktil := \vktil_t(x)$, $\vKtil := \vKtil_t$ and $\vI := \vI_t$.  We first use the definition in \eqref{eq:sigma_update} and the triangle inequality to write
\begin{align}
    &|\sigmatil_t(x)^2 - \sigmaTI_t(x)^2| \nonumber \\
        &~ = \big|\vktil^T (\vKtil + \sigma^2 \vI)^{-1} \vktil  - \vk^T (\vK + \sigma^2 \vI)^{-1} \vk\big| \\
        &~ \le \big| \vktil^T (\vKtil + \sigma^2 \vI)^{-1} \vktil - \vktil^T (\vK + \sigma^2 \vI)^{-1} \vktil \big| + \big| \vktil^T (\vK + \sigma^2 \vI)^{-1} \vktil - \vk^T (\vK + \sigma^2 \vI)^{-1} \vk\big| \label{eq:sigma_triangle} \\
        &:= T_1 + T_2.
\end{align}
We proceed by bounding $T_1$ and $T_2$ separately, starting with the latter.

Set $\vM := (\vK + \sigma^2 \vI)^{-1}$ for brevity. By expanding the quadratic function $(\vktil - \vk)^T  \vM (\vktil - \vk)^T$, grouping the terms appearing in $T_2$, and applying the triangle inequality, we obtain
\begin{equation}
    T_2 \le 2| \vk^T \vM (\vktil - \vk)| + |(\vktil - \vk)^T \vM (\vktil - \vk)|.
\end{equation}
We upper bound each of these terms of the form $\va^T \vM \vb$ by $\|\va\|_2 \|\vM\|_{2\to2} \|\vb\|_2$, where $\|\vM\|_{2\to2}$ is the spectral norm.  By definition, $\lambda$ is an eigenvalue of $\vK$ if and only if $\frac{1}{\lambda + \sigma^2}$ is an eigenvalue of $\vM$; since $\vK$ is positive semi-definite, it follows that $\|\vM\|_{2\to2} \le \frac{1}{\sigma^2}$.  We also have $\|\vk\|_2^2 \le N$ since the entries of $\vk$ lies in $[0,1]$, and $\|\vktil - \vk\|_2^2 \le N^3 \epsilon^2$ since the absolute values of the entries of $\vktil - \vk$ are upper bounded by $N\epsilon$ by the argument following \eqref{eq:AN_def2}.  Combining these, we obtain 
\begin{equation}
    T_2 \le 2\sigma^{-2} N^2\epsilon + \sigma^{-2} N^3 \epsilon^2. \label{eq:T2_bound}
\end{equation}

To bound $T_1$, we use the following inequality for positive definite matrices $\vU,\vV$ and any unitarily invariant norm $\vertiii{\cdot}$ \cite[Lemma X.1.4]{Bha97}:
\begin{equation}
    \vertiii{ (\vU+\vI)^{-1} - (\vU+\vV+\vI)^{-1} } \le \vertiii{ \vI - (\vV+\vI)^{-1} }.
\end{equation}
Specializing to the spectral norm, multiplying through by $\frac{1}{\sigma^2}$, and choosing $\vU = \frac{1}{\sigma^2} \vK$ and $\vV = \frac{1}{\sigma^2}(\vKtil - \vK)$, we obtain
\begin{equation}
    \big\| (\vK+\sigma^2 \vI)^{-1} - (\vKtil + \sigma^2 \vI)^{-1} \big\|_{2\to2} \le \big\| \sigma^{-2} \vI - (\vKtil - \vK + \sigma^2\vI)^{-1} \big\|_{2\to2}. \label{eq:matrix_ineq2}
\end{equation}
Next, $\lambda$ is an eigenvalue of $\vKtil - \vK$ if and only if $\sigma^{-2} - \frac{1}{\lambda+ \sigma^2}$ is an eigenvalue of $\sigma^{-2} \vI - (\vKtil - \vK + \sigma^2\vI)^{-1}$.  Writing $\sigma^{-2} - \frac{1}{\lambda+ \sigma^2} = \sigma^{-2}\big(1 - \frac{1}{\lambda/\sigma^2 + 1}\big) \le \sigma^{-4} \lambda$, it follows that the right-hand side of \eqref{eq:matrix_ineq2} is upper bounded by $\sigma^{-4} \| \vKtil - \vK \|_{2\to2}$.  Using \eqref{eq:frob_bound} (observe that $\vA_N = \vKtil - \vK$) and the fact that the spectral norm is upper bounded by the Frobenius norm, we obtain $\| \vKtil - \vK \|_{2\to2} \le N^2 \epsilon$.  Substituting into \eqref{eq:matrix_ineq2}, we conclude that the matrix $\vM' := (\vK+\sigma^2 \vI)^{-1} - (\vKtil + \sigma^2 \vI)^{-1}$ has a spectral norm which is upper bounded by $\sigma^{-4} N^2 \epsilon$.  Finally, $T_1$ can be written as $\vktil^T \vM' \vktil$, and since $\|\vktil\|_2^2 \le N$ (since each entry of $\vktil$ lies in $[0,1]$), we obtain
\begin{equation}
    T_1 \le \sigma^{-4} N^3 \epsilon. \label{eq:T1_bound}
\end{equation} 
Combining \eqref{eq:T2_bound} and \eqref{eq:T1_bound} and crudely writing $N^2 \epsilon \le N^3 \epsilon$ and $N^3 \epsilon^2 \le N^3 \epsilon$, we obtain
\begin{equation}
    |\sigmatil_t(x)^2 - \sigmaTI_t(x)^2| \le \big( 3\sigma^{-2} + \sigma^{-4} \big) N^3 \epsilon,
\end{equation}
and hence, applying the inequality $(a - b)^2 \le |a^2 - b^2|$, we obtain
\begin{equation}
    \Delta_{t}^{(\sigma)} \le \sqrt{\big( 3\sigma^{-2} + \sigma^{-4} \big) N^3 \epsilon}.
\end{equation}

To characterize $\Delta_{t}^{(\mu)}$, we write the following analog of \eqref{eq:sigma_triangle}:
\begin{align}
    &|\mutil_t(x)^2 - \muTI_t(x)^2| \nonumber \\
        &~ \le \big| \vktil^T (\vKtil + \sigma^2 \vI)^{-1} \vy - \vktil^T (\vK + \sigma^2 \vI)^{-1} \vy \big| + \big| \vktil^T (\vK + \sigma^2 \vI)^{-1} \vy - \vk^T (\vK + \sigma^2 \vI)^{-1} \vy\big| \label{eq:mu_triangle} \\
        &~ := T_1 + T_2.
\end{align}
Following the same arguments as those above, and noting from \eqref{eq:R_assump3} that $\|\vy\|_2^2 \le N \Ltil_N^2$, we obtain
\begin{gather}
    T_1 \le \sigma^{-2} N^2 \epsilon \Ltil_N \\
    T_2 \le \sigma^{-4} N^3 \epsilon \Ltil_N,
\end{gather}
and hence
\begin{equation}
    \Delta_{t}^{(\mu)} \le \big(\sigma^{-2} + \sigma^{-4}\big) N^3 \epsilon \Ltil_N.
\end{equation}

\section{Applications to Specific Kernels (Corollary \ref{cor:specific})}

Throughout this section, we let $\calI_T(z)$ denote the integer in $\{1,\dotsc,T\}$ which is closest to $z \in \real$.  We focus primarily on the proof for TV-GP-UCB, since the proof for R-GP-UCB is essentially identical.

We begin with the squared exponential kernel.  From \cite[Thm.~5]{Sri12}, we have $\gammaNtilTI = \Ord(d\log \Ntil) = \Otil(1)$, and we thus obtain
\begin{equation}
    \bigg(\frac{T}{\Ntil} + 1\bigg) \Big(\gammaNtilTI + N^3 \epsilon\Big) = \Otil\bigg( \Big(\frac{T}{\Ntil} + 1\Big) (1+\Ntil^3 \epsilon) \bigg).
\end{equation}
Setting $\Ntil = \calI_T(\epsilon^{-1/3})$, we find that this behaves as $\Otil(T\epsilon^{1/3})$ when $\epsilon \ge \frac{1}{T^3}$, and as $\Otil(1)$ when  $\epsilon < \frac{1}{T^3}$ (and hence $\Ntil=T$).  Substitution into Theorem \ref{thm:general} yields the desired result.

For the Mat\'ern kernel, we have from \cite[Thm.~5]{Sri12} that $\gammaNtilTI = \Ord(\Ntil^c \log \Ntil) = \Otil(\Ntil^c)$ with $c = \frac{d(d+1)}{2\nu + d(d+1)}$, and we thus obtain 
\begin{equation}
    \bigg(\frac{T}{\Ntil} + 1\bigg) \Big(\gammaNtilTI + \Ntil^3 \epsilon\Big) = \Otil\bigg( \Big(\frac{T}{\Ntil} + 1\Big) (\Ntil^c+\Ntil^3 \epsilon) \bigg).
\end{equation}
Setting $\Ntil = \calI_T(\epsilon^{-\frac{1}{3-c}})$, we find that this behaves as $\Otil(T\epsilon^{\frac{1-c}{3-c}})$ when $\epsilon \ge \frac{1}{T^{3-c}}$, and as $\Otil( T^c)$ when  $\epsilon < \frac{1}{T^{3-c}}$ (and hence $\Ntil=T$).  Substitution into Theorem \ref{thm:general} yields the desired result.

For R-GP-UCB, the arguments are analogous using Theorem \ref{thm:general_R} in place of Theorem \ref{thm:general}, with $N$ playing the role of $\Ntil$.  We set $N = \calI_T(\epsilon^{-1/4})$ for the squared exponential kernel and $N = \calI_T(\epsilon^{-\frac{1}{4-c}})$ for the Mat\'ern kernel.

\section{Lower Bound (Theorem \ref{thm:conv})}

We obtain a lower bound on the regret of \emph{any} algorithm by considering the optimal algorithm for a genie-aided setting.  Specifically, suppose that at time $t$, the entire function $f_{t-1}$ is known perfectly.  We claim that the optimal strategy, in the sense of minimizing the expected regret, is to choose $x_t$ to be any maximizer of $f_{t-1}$.   This can be seen by noting that minimizing the regret $r_t = f_t(x_t^*) - f_t(x_t)$ is equivalent to maximizing the function value $f_t(x_t)$, since $f_t(x_t^*)$ is unaffected by the choice of $x_t$.  Then, conditioned on the entire function $f_{t-1}$, the next value $f_t(x)$ is distributed as $\Ndist(\sqrt{1-\epsilon}f_{t-1}(x) ,\epsilon)$, and clearly the optimal strategy is to choose the point that maximizes the mean.

We proceed by lower bounding the regret incurred by such  a scheme.  Recall that for each $t$, both $f_t$ and $g_t$ are distributed as $\GP(0,k)$.  Thus, \eqref{eq:k_deriv1} and \eqref{eq:k_deriv2} hold for all such functions.

We let $\nabla f$ denote the gradient vector of a function $f$, and let $\nabla^2 f$ denote the Hessian matrix.  For the time being, we condition on the previous function $f_{t-1}$, the selected point $x_t$ (i.e.,~the maximizer of $f_{t-1}$) and the innovation function $g_t$ satisfying the following events for some positive constants $L$ and $\eta$:
\begin{align}
    \calA_1 &:= \bigg\{ \Big| \frac{\partial^2 f_{t-1}(x)}{\partial x^{(j_1)} \partial x^{(j_2)}} \Big| \le L, ~~ \forall j_1,j_2,x \bigg\} \label{eq:A1} \\
    \calA_2 &:= \bigg\{ \Big| \frac{\partial^2 g_t(x)}{\partial x^{(j_1)} \partial x^{(j_2)}} \Big| \le L, ~~ \forall j_1,j_2,x \bigg\} \label{eq:A2} \\
    \calA_3 &:= \bigg\{ \frac{\sqrt{\epsilon} \big| \frac{\partial g_t(x)}{ \partial x^{(j)} } \big| }{2L\sqrt{d}} \le \eta, ~~ \forall j \bigg\} \label{eq:A3} \\
    \calA_4 &:= \big\{ d(x_t,B) \ge \eta \big\}, \label{eq:A4}
\end{align} 
where $d(x_t,B) := \min_{x\in B} \|x_t - x\|_2 $ is the distance of $x_t$ to the closest point on the boundary $B$ of the compact domain $D$.  Observe that for any fixed $\eta$, $\Pr[ \calA_i ]$ can be made arbitrarily close to one for $i=1,2,3$ by choosing $L$ sufficiently large. Moreover, we have $\Pr[\calA_4] > 0$ for sufficiently small $\eta$, since otherwise the maximum of $f \sim \GP(0,k)$ would be on the boundary of the domain $D$ with probability one.  Applying the union bound, we conclude that the event $\calA := \calA_1 \cap \calA_2 \cap \calA_3 \cap \calA_4$ occurs with strictly positive probability for suitable chosen $\eta$ and $L$.
 
We fix an arbitrary vector $v$ with $\|v\|_2 = 1$ and a constant $\delta > 0$, and note that the regret $r_t$ at time $t$ an be lower bounded as follows provided that $x_t + v\delta \in D$:
\begin{align}
    r_t &= \max_{x} f_t(x) - f_t(x_t)  \\
         &\ge f_t(x_t+v\delta) - f_t(x_t) \\
         &= \sqrt{1-\epsilon}\big( f_{t-1}(x_t + v\delta) - f_{t-1}(x_t) \big) + \sqrt{\epsilon}\big( g_t(x_t+v\delta) - g_t(x_t) \big) \label{eq:regretLB3} \\
         &= \sqrt{1-\epsilon}\,\frac{1}{2}\delta^2 v^T \big[ \nabla^2 f_{t-1}(x_t+v\delta_f) \big] v + \sqrt{\epsilon}\big( \delta v^T \nabla g_t (x_t) + \frac{1}{2}\delta^2 v^T \big[ \nabla^2 g_t(x_t+v\delta_g) \big] v\big), \label{eq:regretLB4}
\end{align}
where \eqref{eq:regretLB3} follows by substituting the update equations in \eqref{eq:tv_model0}--\eqref{eq:tv_model}, and \eqref{eq:regretLB4} holds for some $\delta_f \in [0, \delta]$ and $\delta_g\in [0,\delta]$ by a second-order Taylor expansion; note that $\nabla f_{t-1}(x_t) = 0$ since $x_t$ maximizes $f_{t-1}$, whose peak is away from the boundary of the domain by \eqref{eq:A4}.

We choose the unit vector $v$ to have the same direction as $\nabla g_t(x_t)$, so that $\delta v^T \nabla g_t(x_t) = \delta \|\nabla g_t(x_t) \|_2$. By \eqref{eq:A1}--\eqref{eq:A2}, the entries of $\nabla^2 f_{t-1}(x_t+v\delta_f)$ and $\nabla^2 g_t(x_t+v\delta_g) $ are upper bounded by $L$, and thus a standard inequality between the entry-wise $\ell_\infty$ norm and the spectral norm reveals that the latter is upper bounded by $Ld$.  This, in turn, implies that $v^T \big[ \nabla^2 f_{t-1}(x_t+v\delta_f) \big] v$ and $ v^T \big[ \nabla^2 g_t(x_t+v\delta_g) \big] v$ are upper bounded by $Ld$, and hence
\begin{align}
    r_t &\ge \sqrt{\epsilon} \delta \|\nabla g_t(x_t) \|_2 - \frac{1}{2} Ld\delta^2 \big( \sqrt{1+\epsilon} + \sqrt{\epsilon} \big) \\
         &\ge  \sqrt{\epsilon} \delta \|\nabla g_t(x_t) \|_2 - Ld\delta^2,
\end{align}
where we have used $ \sqrt{1+\epsilon} + \sqrt{\epsilon} \le 2$.  By differentiating with respect to $\delta$, it is easily verified that the right-hand side is maximized by $\delta = \frac{\sqrt{\epsilon} \|\nabla g_t(x_t) \|_2 }{ 2Ld }$.  This choice is seen to be valid (i.e.,~it yields $x_t + v\delta$ still in the domain) by \eqref{eq:A3}--\eqref{eq:A4} and the fact that $\|z\|_2 \le \sqrt{d}\|z\|_{\infty}$ for $z \in \real^d$, and we obtain
\begin{equation}
    r_t \ge \frac{\epsilon \|\nabla g_t(x_t) \|_2^2 }{ 4Ld }.
\end{equation}
It follows that the expectation of $r_t$ is lower bounded by
\begin{align}
    \myexpect[r_t]
         &\ge \Pr[\calA] \myexpect[r_t | \calA] \label{eq:LB_end1} \\
         &\ge \Pr[\calA] \frac{\epsilon\, \myexpect\big[ \|\nabla g_t(x_t) \|_2^2 \,|\, \calA \big] }{ 4Ld }  \\
         &= \Theta(\epsilon), \label{eq:LB_end3}
\end{align}
where \eqref{eq:LB_end1} follows since $r_t \ge 0$ almost surely, and \eqref{eq:LB_end3} follows since $\myexpect\big[ \|\nabla g_t(x_t) \|_2^2 \,|\, \calA \big] >0$ by a simple proof by contradiction: The expectation can only equal zero if its (non-negative) argument is zero almost surely, but if that were the case then the unconditional distribution of $\|\nabla g_t(x_t) \|_2^2$ would satisfy $\Pr[ \|\nabla g_t(x_t) \|_2^2 = 0] > \Pr[\calA]$, which is impossible since the entries of $\nabla g_t(x_t) $ are Gaussian \cite[Sec.~9.4]{Ras06} and hence have zero probability of being exactly zero.

Finally, using \eqref{eq:LB_end3}, the average cumulative regret satisfies $\myexpect[R_T] = \sum_{i=1}^T \myexpect[r_T] = \Omega(T\epsilon)$.

\section{Further results for traffic speed data} \label{sec:TRAFFIC_RESULTS}
\begin{figure*}[h]
\centering
\setcounter{subfigure}{0}
 \begin{subfigure}[b]{0.24\textwidth}
	\includegraphics[width=\textwidth]{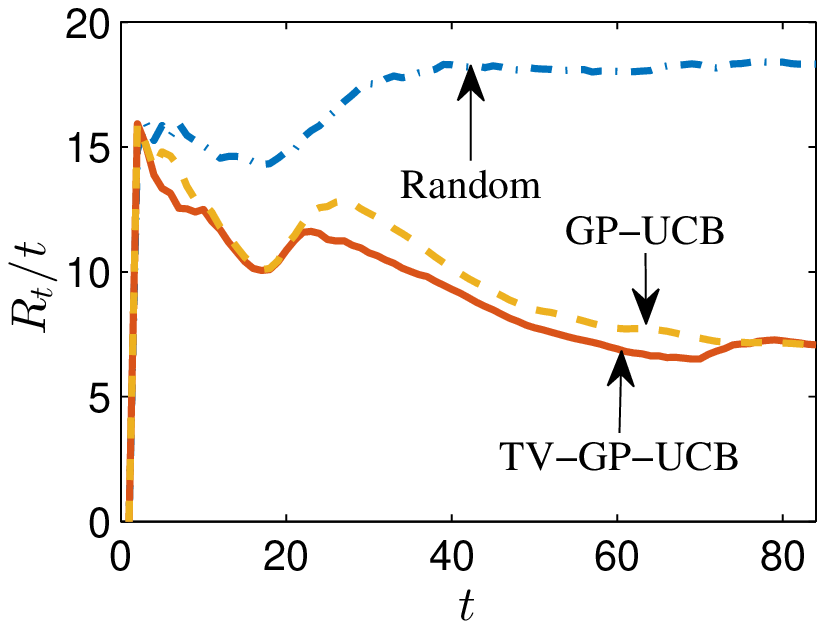}  
	\caption{Day 1}
	\label{fig:td_day1a}
  \end{subfigure}
\setcounter{subfigure}{1}
  \begin{subfigure}[b]{0.24\textwidth}
    \includegraphics[width=\textwidth]{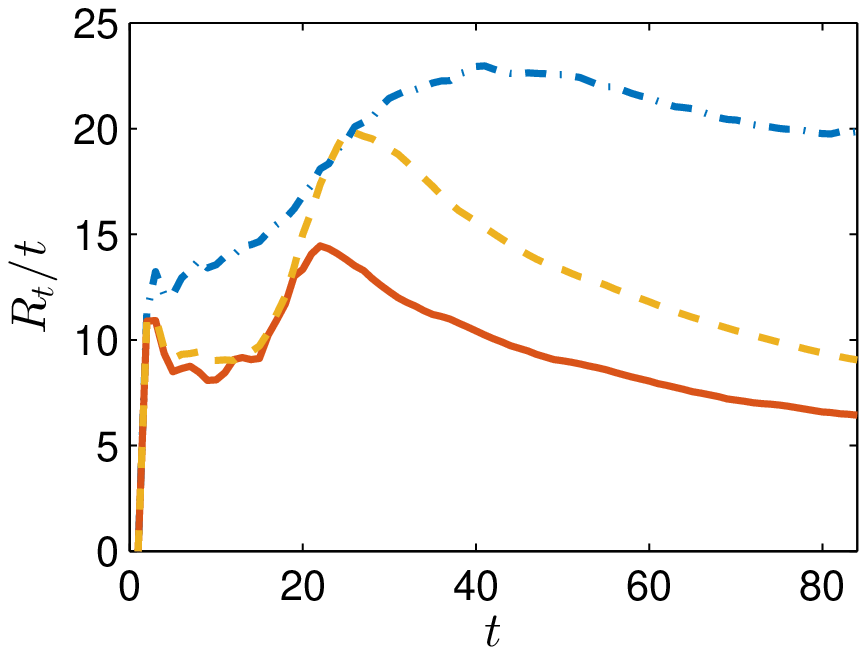}
	\caption{Day 2}
	\label{fig:td_day2a}  
  \end{subfigure}
\setcounter{subfigure}{2}
	\begin{subfigure}[b]{0.24\textwidth}
	 \includegraphics[width=\textwidth]{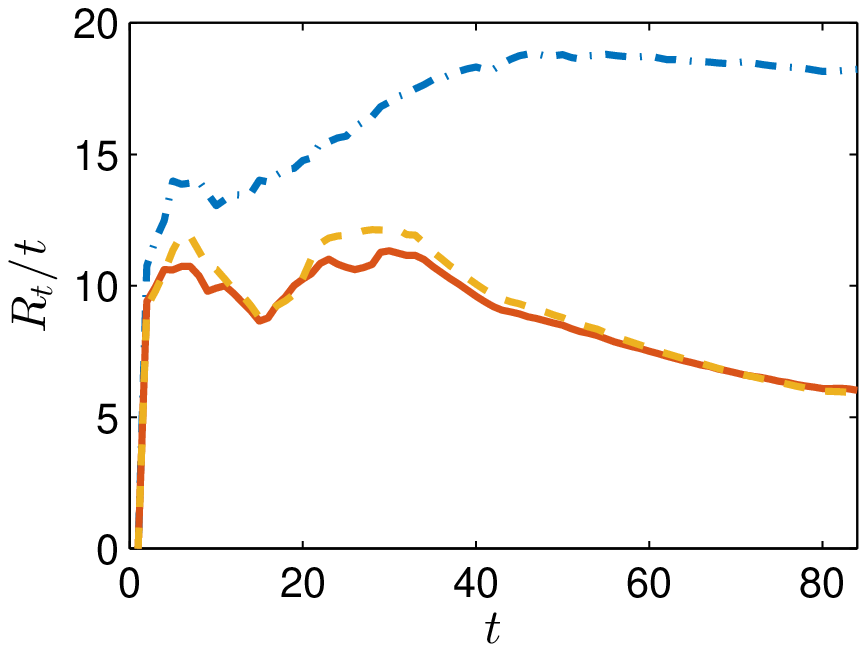}  
	\caption{Day 3}
	\label{fig:td_day3a}
  \end{subfigure}
\setcounter{subfigure}{3}
	\begin{subfigure}[b]{0.24\textwidth}
	 \includegraphics[width=\textwidth]{figures/final_figs/Jon2/day4.eps}  
	\caption{Day 4}
	\label{fig:td_day4a}
  \end{subfigure}
\setcounter{subfigure}{4}
  \begin{subfigure}[b]{0.24\textwidth}
	\includegraphics[width=\textwidth]{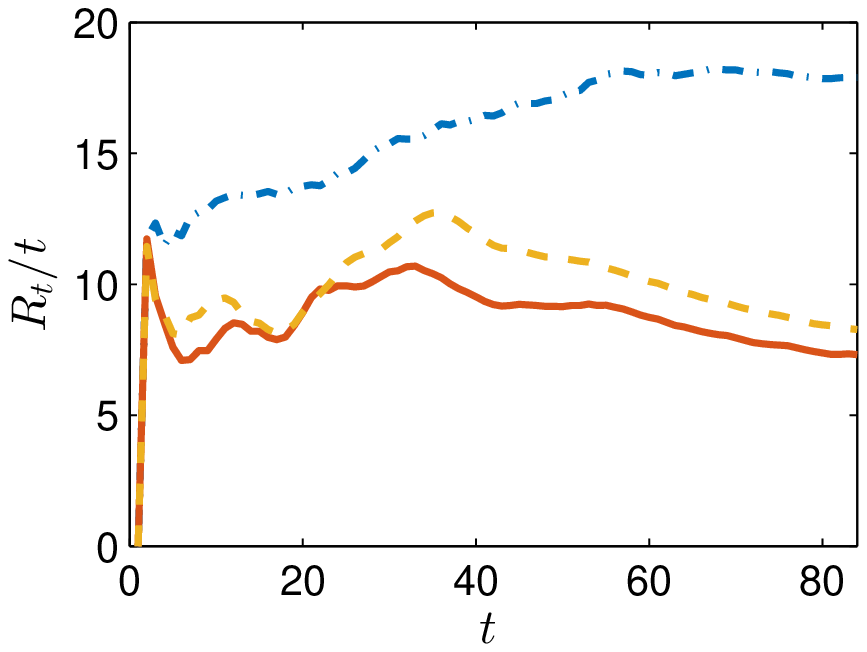}
	\caption{Day 5}  
	\label{fig:td_day5a}	 
  \end{subfigure}
\setcounter{subfigure}{5}
  \begin{subfigure}[b]{0.24\textwidth}
   	\includegraphics[width=\textwidth]{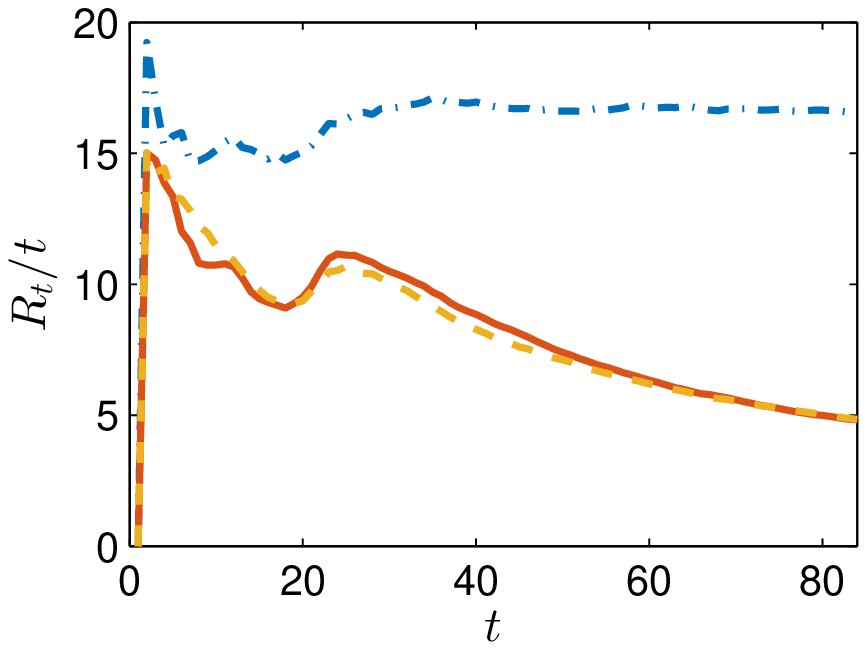}
	\caption{Day 6}
	\label{fig:td_day6a} 
  \end{subfigure} 
\setcounter{subfigure}{6}
  \begin{subfigure}[b]{0.24\textwidth}
	 \includegraphics[width=\textwidth]{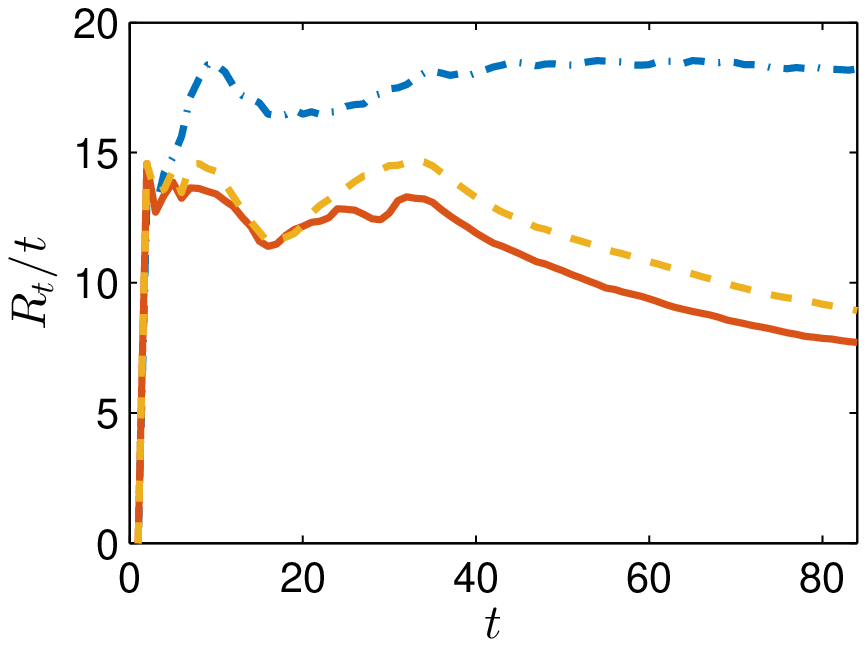}  
	\caption{Day 7}
	\label{fig:td_day7a}
  \end{subfigure} 
 \setcounter{subfigure}{7}
  \begin{subfigure}[b]{0.24\textwidth}
	 \includegraphics[width=\textwidth]{figures/final_figs/Jon2/day8.eps}  
	\caption{Day 8}
	\label{fig:td_day8a}
  \end{subfigure} 
\setcounter{subfigure}{8}
  \begin{subfigure}[b]{0.24\textwidth}
	 \includegraphics[width=\textwidth]{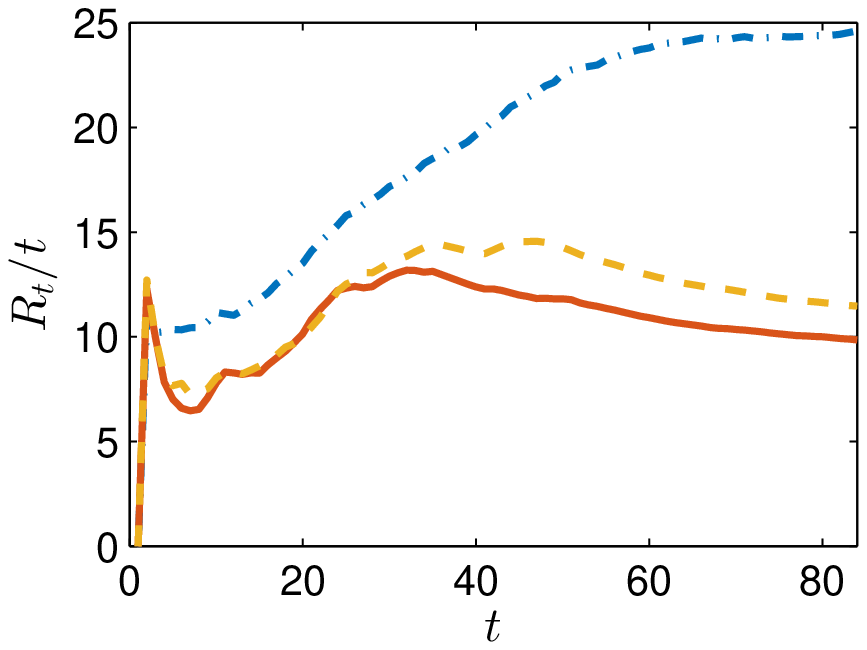}  
	\caption{Day 9}
	\label{fig:td_day9a}
  \end{subfigure} 
\setcounter{subfigure}{9}
  \begin{subfigure}[b]{0.24\textwidth}
	 \includegraphics[width=\textwidth]{figures/final_figs/Jon2/day10.eps}  
	\caption{Day 10}
	\label{fig:td_day10a}
  \end{subfigure} 
  \caption{Numerical performance of upper confidence bound algorithms on traffic speed dataset.}
  \label{fig:appendix}
\end{figure*}

In Figure~\ref{fig:appendix}, we outline the complete results on all the testing days for the experiment described in Section~\ref{r_s_data}. The sensors used in the experiment have the following IDs:
[0, 54, 69, 77, 169, 131, 262, 216, 34, 320, 308, 177, 130, 221, 290, 348, 25, 157, 252, 83, 163, 149, 294, 21, 246, 45, 98, 74, 274, 237, 322, 29, 120, 44, 49, 241, 286, 99, 247, 297, 96, 234, 236, 205, 329, 214, 28, 175, 65, 220].

\end{document}